%% file: main.tex
\documentclass[10pt,twocolumn,letterpaper]{article}

\usepackage[pagenumbers]{wacv} % PREPRINT: camera-ready layout with author block and page numbers

\input{preamble}

\definecolor{wacvblue}{rgb}{0.21,0.49,0.74}
\usepackage[breaklinks,colorlinks,allcolors=wacvblue]{hyperref} % no pagebackref in the preprint
\usepackage{microtype}
\hypersetup{
  pdftitle={GuidedWAM: Training-Free Energy Guidance for World Action Models in Autonomous Driving},
  pdfauthor={Xiyan Su, Finn Rasmus Schäfer, Hojun Lim, Frank Diermeyer, Johannes Betz, Markus Lienkamp},
  pdfsubject={Preprint. Accepted at the WAM-AD workshop at WACV 2027},
  pdfkeywords={world action models, energy guidance, flow matching, end-to-end autonomous driving, NAVSIM}
}

\newcommand{\pos}[1]{\left[#1\right]_{+}} % positive part, used in the appendix

\def\wacvPaperID{*****} % *** Enter the WACV Paper ID here
\def\confName{WACV}
\def\confYear{2027}

\title{GuidedWAM: Training-Free Energy Guidance for World Action Models in Autonomous Driving}

\author{%
Xiyan Su$^{1,3}$\thanks{Corresponding author: \texttt{xiyan.su@tum.de}} \quad
Finn Rasmus Sch\"afer$^{2,3}$ \quad
Hojun Lim$^{1,3}$\\
Frank Diermeyer$^{1,3}$ \quad
Johannes Betz$^{2,3}$ \quad
Markus Lienkamp$^{1,3}$\\[4pt]
$^{1}$Institute of Automotive Technology (FTM), Technical University of Munich\\
$^{2}$Professorship Autonomous Vehicle Systems (AVS), Technical University of Munich\\
$^{3}$Munich Institute of Robotics and Machine Intelligence (MIRMI), Technical University of Munich\\[4pt]
Project page: \url{https://tumftm.github.io/GuidedWAM/}
}

\begin{document}
\maketitle
% IEEE copyright notice for the accepted version, as an unnumbered footnote on the first page
\begingroup
\renewcommand{\thefootnote}{}%
\footnotetext{\hspace{-1.8em}Accepted at the WAM-AD workshop at WACV 2027. \copyright~2027 IEEE. Personal use of this material is permitted. Permission from IEEE must be obtained for all other uses, in any current or future media, including reprinting/republishing this material for advertising or promotional purposes, creating new collective works, for resale or redistribution to servers or lists, or reuse of any copyrighted component of this work in other works.}%
\endgroup
\input{sections/00_abstract}
\input{sections/01_intro}
\input{sections/02_related_works}
\input{sections/03_method}

\input{sections/04_results}
\input{sections/05_conclusion}
\input{sections/99_acknowledgement}
{
    \small
    \bibliographystyle{ieeenat_fullname}
    \bibliography{references}
}

\input{sections/A_supplementary}

\end{document}

%% file: preamble.tex
\newcommand{\sd}[1]{{\color{gray}\tiny$\pm$#1}}
\newcommand{\nosd}{\phantom{\sd{0.0}}}

\usepackage{multirow}
\usepackage{algorithm}
\usepackage{algpseudocode}

%% file: sections/00_abstract.tex
\begin{abstract}
World action models (WAMs) for autonomous driving couple the driving policy and the visual world model in a single generative network whose objectives are fixed post-training. 
Conditioning on a new driving objective requires retraining. 
We study energy guidance as a general mechanism and present \textbf{GuidedWAM}, a model-agnostic framework for injecting driving objectives into frozen WAMs at sampling time. 
Our framework constructs each objective as differentiable energies on the sampler's intermediate clean-action estimate, which needs only forward access to the frozen model. 
On the NAVSIM v2 hard-split, guidance improves all three frozen, publicly released WAMs by $2.2$ to $3.8$ EPDMS over multi-seed experiment, and adds less than $10\%$ to the sampling time per step.
These results position energy guidance as a practical post-training control channel for WAMs.
\end{abstract}

%% file: sections/01_intro.tex
\section{Introduction}
\label{sec:intro}

World action models (WAMs) are an emerging family of end-to-end driving models.
The generative model learns the driving policy simultaneously with a visual world model, where the policy inherits the physical and visual knowledge of a large video prior, and the model can imagine the future driving scene \citep{drivewam2026,simwam2026,universe2026,xia2026drivelaw}.
Recent WAMs reach planning scores above 90 PDMS on NAVSIM v1 \citep{dauner2024navsim,drivewam2026,simwam2026}.
% and achieve real-time deployment by turning off the imagination at inference \citep{yuan2026fastwam,ma2026fasterwam}.

Despite these advantages, WAMs come with their own drawbacks. By training as a whole, the driving objectives freeze once the training is finished and cannot be extended at will. 
In practice, a driving policy has to serve objectives that change with the operating context during deployment: a wider margin to the road edge on a narrow street, a larger clearance around vulnerable road users, a temporary speed limit, or a passenger's comfort preference.
A WAM learns its driving behavior mainly from driving data. 
Sampling-time conditioning is only possible if such driving objectives are available in the data during training.
Adding one means collecting supervision and retraining a billion-parameter model, which has to be repeated for every new objective.
What is missing is a control channel that acts on the frozen model at sampling time.

Therefore, we present \textbf{GuidedWAM}, a training-free energy-guidance framework for world action models for autonomous driving.
Each driving objective is constructed as a differentiable energy on the sampler's one-step clean estimate of the action stream, whose gradient is added to the frozen model's velocity field during sampling.
We investigate whether the driving policy of a frozen WAM can be guided at sampling time toward an objective it was never trained on.
% We investigate two questions raised by this control channel separately. \textbf{RQ1} Can the driving policy of a frozen WAM be guided at sampling time toward an objective it was never trained on? \textbf{RQ2} Does the visual imagination follow the guided policy even when it is not explicitly guided?

% For \textbf{RQ1}, we inject a new objective for drivable area into three state-of-the-art open-source WAMs on the NAVSIM v2, and we show that energy guidance improves all three, by $1.5$ to $4.8$ EPDMS points.
% % Check this number again for both oracle and non-oracle 
% Although the additional control channel helps every model, the optimal guidance strength differs widely from one model to another, so the strength has to be calibrated for each model before deployment.

% For \textbf{RQ2}, we run a controlled case study that compares a joint WAM with an action-conditioned video world model.
% Both are built from the same video backbone and trained with the same data, recipe, and budget.
% The action-conditioned model serves as the reference for the condition channel as the generated video is conditioned on the given plan.
% We show that counterfactual-plan training, applied on a fraction of the pre-training steps, makes the video generation of the WAM follow the guided policy \pending{joint WAM result}, and we benchmark its action controllability against the action-conditioned reference on ACT-Bench \pending{ACT-Bench numbers for both models}.

\noindent Our contributions are \textbf{two-fold}:
\begin{itemize}
  \item We derive energy guidance on the \textit{one-step clean estimate} of the flow-matching sampler that current WAMs share, in a \textit{Jacobian-free} form that needs only forward passes through the frozen model, and show that it steers three public WAMs on NAVSIM v2.
  \item We show that the guidance is stabilized by \textit{sampling-time guidance scheduling} and a \textit{velocity-bounded correction} through an ablation study.
  % \item We show that counterfactual-plan training during the pre-training phase makes the visual imagination follow the guided policy later at sampling time, and benchmark the controllability on ACT-Bench.
\end{itemize}

%% file: sections/02_related_works.tex
\section{Related Works}
\label{sec:related}

\textbf{World action models as driving planners}
A video world model learns how an environment evolves under the agent's actions, so that a policy can be trained or evaluated on imagined rollouts \citep{ha2018worldmodels,hafner2020dreamer}.
In autonomous driving, this idea first produced action-conditioned video world models, which render the future camera view from past frames and actions \citep{wang2023drivedreamer,hu2023gaia,gao2024vista}.
World action models close the gap by also generating the action stream, so that the driving knowledge learned from a large video prior can transfer to planning.
DriveWAM trains video and ego action tokens as one token sequence under a joint flow-matching objective \citep{drivewam2026}, whereas UNIVERSE shares one transformer between the two modalities and decouples them with an attention mask \citep{universe2026}. SimWAM co-trains a video expert with a light action expert and drops the video branch at inference for run-time \citep{simwam2026}, and DriveLaW plans from the latent of its video generator \citep{xia2026drivelaw}.
For this paper, the three public models we guide share one property: the plan is produced by a flow-matching sampler conditioned on a fixed set of context channels, past frames, ego status, and a navigation command.\\
\textbf{Control channels for flow-matching planners}
Most current WAMs are built on diffusion or flow-matching backbones \citep{lipman2023flow,liu2023rectified,peebles2023dit}.
Conditional flow matching trains the velocity field on a context.
Classifier guidance adds the gradient of a noise-aware classifier to the score at sampling time \citep{dhariwal2021classifier}, and classifier-free guidance replaces the classifier with the model's own unconditional prediction \citep{ho2022cfg}.
Independent of training, energy guidance instead descends an external differentiable objective on the sampler's clean estimate. Diffusion posterior sampling evaluates the objective on the denoiser's clean estimate \citep{chung2023dps}. Manifold-preserving guided diffusion drops the network Jacobian \citep{he2024mpgd}. \citet{feng2025flowguidance} carry the construction to flow matching.
In the driving domain, CTG and CTG++ guide traffic simulation with signal-temporal-logic and language-derived costs \citep{zhong2023ctg,zhong2023ctgpp}. Diffusion Planner applies classifier guidance to a vectorized-state planner \citep{zheng2025diffusionplanner}. GuideFlow corrects constraint violations inside the sampling loop with a constraint vocabulary fixed at training \citep{guideflow2025}.
A second family steers a frozen planner from outside the sampler by search rather than by gradients.
Diffusion-ES applies evolutionary search to a trajectory diffusion model \citep{yang2024diffusiones}. TOAD runs cross-entropy-method search over the proposals of $7$ frozen NAVSIM planners against a learned trajectory scorer \citep{xu2026toad}.
We adopt the Jacobian-free clean-estimate gradient.
We apply the analytic energies directly on the diffusion policy instead of searching or scoring sampled trajectories.\\
\textbf{Safety filters and constrained flow matching}
Another way to enforce a driving constraint on a learned planner is a safety filter applied after the plan is produced, a control barrier function certificate that defines a safe set and a minimal correction that keeps the plan inside it \citep{ames2019cbf}.
SafeFlow enforces control barrier functions across the sampler \citep{dai2025safeflow}, SafeFlowMatcher alternates prediction with barrier-based correction \citep{yang2025safeflowmatcher}, UniConFlow shapes a prescribed-time guidance field for equality and inequality constraints \citep{yang2025uniconflow}, HardFlow solves a receding-horizon optimization on the clean estimate at every step \citep{li2025hardflow}.
Safety-guided flow shows that negative guidance must act within a critical time window \citep{kim2026sgf}, which is the closest precedent for our noise-level scheduling.
These methods target hard satisfaction of known constraint sets on planners.
We construct soft and composable energies for frozen models and compare the in-sampling channel with the post-hoc member of this lineage on the same constraint.

% Our objectives are soft and composable, the models are frozen, and we report a measured operating range rather than a guarantee.
% We therefore do not position guidance against the filter but next to it: we compare the in-sampling channel with the post-hoc member of this family on the same constraint, and we ask what each channel moves that the other cannot.
% Our objectives are soft and composable, the models are frozen, and we report a measured operating range rather than a guarantee. A filter also acts only where the constraint is violated, so its corrections are local to the violation, and without a dynamics model that locality appears as a kink and a comfort cost, whereas guidance shapes the whole plan before the violation occurs. We therefore do not position guidance against the filter but next to it: we compare the in-sampling channel with the post-hoc member of this family on the same constraint and ask what each channel moves that the other cannot.

%% file: sections/03_method.tex
\section{Method}
\label{sec:method}

\subsection{Preliminaries}
\label{sec:prelim}

Flow matching learns a time-dependent velocity field that transports a simple prior to the data distribution \citep{lipman2023flow, liu2023rectified}.
We consider a frozen WAM whose action stream $\tau \in \mathbb{R}^{T \times d}$, an ego trajectory or an action-token sequence, is generated by such a sampler conditioned on the observation context $X$.
Along the Gaussian flow-matching path, the noisy state is the linear interpolation $\tau_t = (1-t)\,\tau_0 + t\,\varepsilon$ with $\varepsilon \sim \mathcal{N}(0, I)$, the network predicts the velocity $v_\theta(\tau_t, t, X) \approx \mathbb{E}[\varepsilon - \tau_0 \mid \tau_t, X]$, and sampling integrates $\tau \leftarrow \tau - \Delta t\, v_\theta$ from $t{=}1$ down to $t{=}0$.
All three public WAMs we guide use this sampler.
Guidance mechanisms are statements about the score $s_t = \nabla_{\tau_t} \log p_t(\tau_t \mid X)$. On a Gaussian flow-matching path, score and velocity are interchangeable through an affine identity \citep{zheng2023guidedflows}.
A short calculation gives
\begin{equation}
v(\tau_t, t) \;=\; -\,\frac{\tau_t + t\, s_t(\tau_t)}{1 - t}
\label{eq:sv}
\end{equation}
so an additive score correction $\Delta s$ enters the velocity scaled by $-t/(1-t)$.
We therefore derive in score space and implement in the velocity space that every sampler exposes.

\subsection{Energy Guidance on One-Step Clean Estimation of Flow-Matching}
\label{sec:derivation}

\textbf{One-step clean estimate}
For every flow-matching sampler, the one-step estimate of the clean sample at each step is the posterior mean under the model, similar to \citet{chung2023dps},
\begin{equation}
\hat\tau_0(\tau_t) \;=\; \tau_t - t\, v_\theta(\tau_t, t, X) \;\approx\; \mathbb{E}[\tau_0 \mid \tau_t, X]
\label{eq:x0hat}
\end{equation}
The driving objectives can be evaluated on this one-step clean estimate.\\
\textbf{Differentiable energies on the clean estimate}
A driving objective is a differentiable energy $\mathcal{E}: \mathbb{R}^{T \times d} \to \mathbb{R}$ on the clean action sequence.
In practice the clean estimate is first decoded to ego trajectory with position and heading at each waypoint. The energy is derived analytically on it: a distance to a lane centerline, a signed distance of the vehicle to the drivable boundary, or a temporary speed limit. \\
% The energies used in this paper are given in Appendix~\ref{app:impl}.
% \textbf{Guided score}
% The target of guidance is the tilted posterior
% \begin{equation}
% p_\gamma(\tau_0 \mid X) \;\propto\; p(\tau_0 \mid X)\, e^{-\gamma E(\tau_0)},
% \label{eq:tilted}
% \end{equation}
% the frozen prior reweighted toward low energy.
% Introducing an optimality variable $\mathcal{O}$ with $p(\mathcal{O} \mid \tau_0) \propto e^{-\gamma E(\tau_0)}$ turns Eq.~\eqref{eq:tilted} into a conditioning problem, and Bayes' rule gives the guided score at every noise level without any training:
% \begin{equation}
% s_t^{\gamma}(\tau_t) \;=\; s_t(\tau_t) \;+\;
% \nabla_{\tau_t} \log \mathbb{E}_{p(\tau_0 \mid \tau_t, X)} \big[e^{-\gamma E(\tau_0)}\big].
% \label{eq:bayes}
% \end{equation}
% The expectation over the denoising posterior is intractable, and the approximation chain that follows is where every practical method, including ours, lives. \\
\textbf{Guided score}
The frozen model draws plans from $p(\tau_0 \mid X)$, the posterior over the action stream given the observation context.
Every objective that shaped this posterior was fixed when training ended.
Guidance changes the target distribution without changing the model. We want plans drawn from
\begin{equation}
p_\gamma(\tau_0 \mid X) \;\propto\; p(\tau_0 \mid X)\, e^{-\gamma \mathcal{E}(\tau_0)}
\label{eq:tilted}
\end{equation}
the same posterior tilted toward low energy by a strength $\gamma > 0$.
% When several objectives are declared, $\mathcal{E}$ is their weighted sum, so the tilt is a composite energy field over the trajectory and the objectives compose by addition.
To sample from Eq.~\eqref{eq:tilted} with a sampler built for $p(\tau_0 \mid X)$, we write the tilt as conditioning.
We denote the driving objective as an auxiliary variable $\mathcal{G}$ with likelihood $p(\mathcal{G} \mid \tau_0) \propto e^{-\gamma \mathcal{E}(\tau_0)}$, denoting the graded preference for low-energy.
Since $\mathcal{G}$ depends on $X$ only through $\tau_0$, Bayes' rule gives $p(\tau_0 \mid \mathcal{G}, X) = p_\gamma(\tau_0 \mid X)$, so sampling the tilted posterior is sampling the frozen model conditioned on $\mathcal{G}$, and the score of the conditioned marginal follows without any training:
\begin{equation}
s_t^{\gamma}(\tau_t) \;=\; s_t(\tau_t) \;+\;
\nabla_{\tau_t} \log \mathbb{E}_{p(\tau_0 \mid \tau_t, X)} \big[e^{-\gamma \mathcal{E}(\tau_0)}\big]
\label{eq:bayes}
\end{equation}
However, the expectation over the denoising posterior is intractable. \\
\textbf{Energy as a gradient}
Replacing the average over the denoising posterior by the model's clean estimate, as in diffusion posterior sampling \citep{chung2023dps}, makes the log-expectation $-\gamma \mathcal{E}(\hat\tau_0(\tau_t))$, whose gradient by the chain rule is
\begin{equation}
\nabla_{\tau_t}\, \mathcal{E}\big(\hat\tau_0(\tau_t)\big) \;=\; J_t^{\top}\, \nabla_{\hat\tau_0} \mathcal{E}(\hat\tau_0)
\label{eq:jacobian}
\end{equation}
where $J_t = \partial \hat\tau_0 / \partial \tau_t = I - t\, \partial v_\theta / \partial \tau_t$ is the Jacobian of the clean estimate, which contains the network Jacobian.
However, as $J_t$ contains the network Jacobian, evaluating it requires a full backward pass through the frozen denoiser at every sampling step, with the compute and activation memory of a training step.
At the scale of a billion-parameter video WAM this is impractical.
We replace $J_t$ with the identity and keep only the energy's own gradient, evaluated at the detached clean estimate, the shortcut of manifold-preserving guided diffusion \citep{he2024mpgd}.
Mapping the score correction into velocity space through Eq.~\eqref{eq:sv} yields the update we run:
\begin{equation}
v' \;=\; v_\theta \;+\; \lambda(t)\, \nabla_{\hat\tau_0} \mathcal{E}(\hat\tau_0)
\qquad
\lambda^{*}(t) \;=\; \gamma\,\frac{t}{1-t}
\label{eq:jf}
\end{equation}
where $\lambda^{*}$ is the canonical schedule implied by the exact score-velocity correspondence.\\
\textbf{Differentiable without backpropagation}
Because the clean estimate is detached, the only gradient ever taken is that of the analytic energy with respect to a $T \times d$ tensor.
The frozen model is run forward once per step, exactly as in unguided sampling with no activation stored.
This is what makes the channel usable on released checkpoints, as it needs forward access to the model and nothing else, and its per-step cost is close to unguided sampling.
The difference is that the deleted $J_t$ acted as a preconditioner that projected the push onto locally plausible directions.
Dropping it increases the chances of off-manifold corrections. 
The canonical $\lambda^{*}$ diverges as $t \to 1$, exactly where the clean estimate is the least reliable.
Both effects are handled by the stabilization methods in \S\ref{sec:stack}.
We show the Jacobian-free form in Eq.~\eqref{eq:jf} can be stabilized empirically for steering frozen flow-matching WAMs within a measurable operating range.

\subsection{Guidance Stabilization}
\label{sec:stack}

Applying Eq.~\eqref{eq:jf} naively can fail for various reasons. A schedule over the noise level and a bound on the correction stabilize the guidance.
Let $\mathcal{E}_1, \dots, \mathcal{E}_K$ be the composed energies with weights $w_k$, and $\Delta v$ be the velocity update in Eq.~\eqref{eq:jf} in the form of $v' \;=\; v_\theta \;+\; \lambda(t) \Delta v$.\\
\textbf{Sampling-time guidance scheduling}
The canonical guidance $\lambda^{*}(t)$ in Eq.~\eqref{eq:jf} places the largest correction at high noise, i.e., $t \to 1$, where the clean estimate is a coarse average of futures and its gradient carries little information. On the other hand, a constant guidance where $\lambda = const$ corrects every step equally, where the early corrections can push the noisy state off the manifold of plausible trajectories and continue to schedule high guidance at $t \to 0$, where the plan is fixed.
% Figure~\ref{fig:schedule} traces the clean estimate through the sampler on one scene: with a uniform weight the estimate leaves the spread of the stock sampler from the middle of the schedule, whereas the trapezoid keeps it inside that spread until the window opens and lands the correction in the final steps.
% \begin{figure*}[t]
% \centering
% \includegraphics[width=\textwidth]{figures/fig_x0_trace.png}
% \caption{Where in the sampler the correction lands, on one NAVSIM v2 hard-split scene of DriveWAM. Left: the one-step clean estimate $\hat\tau_0$ at every denoising step, coloured by denoising progress, for the stock sampler, uniform guidance at $\lambda{=}8$ on every step, and the trapezoidal schedule with $\lambda_{\max}{=}4$. Right: mean lateral offset of the clean estimate from the stock sampler's clean estimate at each step. The grey band is the spread of the stock sampler over $16$ seeds, and the dotted lines give the guidance weight of each schedule on the right axis. Uniform guidance moves the clean estimate away from the stock envelope from the middle of the schedule, where the estimate is still a coarse average of futures. The trapezoid leaves the estimate inside the stock envelope until the window and applies the correction in the last steps, where the plan is settled.}
% \label{fig:schedule}
% \end{figure*}
We therefore truncate the schedule to a trapezoidal window over the noise level. The window reaches maximal over its middle, and is linearly ramped on both sides.
The correction thus acts only once the clean estimate is informative and stops before the last steps, which are left to the frozen model to bring the plan back onto its manifold. \\
\noindent\textbf{Composition}
Several energies are combined as a weighted sum of their gradients,
\begin{equation}
\Delta v \;=\; \sum_{k=1}^{K} w_k\, \nabla_{\hat\tau_0} \mathcal{E}_k(\hat\tau_0)
\label{eq:norm}
\end{equation}
where $\lambda(t)$ carries the overall strength and the fixed relative weights $w_k$ account for the different scales of the energies. \\
% The weights in Table~\ref{tab:combo} and the relative correction reported in Section~4 both use this convention. \\
\textbf{Velocity-bounded correction}
The applied correction is bounded relative to the model's own velocity at the same step,
\begin{equation}
\Delta v \;\leftarrow\; \min\!\Big(1,\; \frac{c\,\lVert v_\theta \rVert_2}{\lambda(t)\,\lVert \Delta v \rVert_2}\Big)\, \Delta v
\label{eq:clip}
\end{equation}
so that $\lambda(t)\,\lVert \Delta v \rVert_2 \le c\,\lVert v_\theta \rVert_2$ at every step, with the ratio $c = 12$ shared across models.
The bound is a safeguard against divergence.
It is not reached at the operating points and limits the correction only at strengths far above them (\S\ref{sec:ablation}).

%% file: sections/04_results.tex
\section{Experiments}

\subsection{Setups}

\paragraph{Dataset and metrics}
We evaluate the planning policy performance on the NAVSIM v2 hard-split \citep{cao2025navsimv2}, which pairs $450$ real scenes with $5{,}462$ synthetic scenes rendered from perturbed initial poses.
We report the extended predictive driver model score (EPDMS) in points, together with its real-scene (S1) and synthetic-scene (S2) stage scores as well as the sub-metric scores.

\subsection{Planning policy performance}
\label{sec:planning}

\paragraph{Stock vs guidance on three frozen WAMs}
Table~\ref{tab:main} reports the three public WAMs with and without the energy guidance (EG) on NAVSIM v2 hard-split. 
Due to the restriction of allowed information from the benchmark, we investigate the performance improvement with one camera-based drivable area energy. 
A combination of energy guidance with additional privileged information is investigated in \S\ref{sec:ablation}.
The drivable area energy reads a camera-only field, a signed distance to the road edge regressed from the front cameras.
This field is predicted by a separate network that is trained once on the training split of the benchmark. The weights of the WAMs are never updated.
Energy guidance improves all three, by $2.3$ points on DriveWAM, $2.2$ on SimWAM and $3.8$ on DriveLaW, since DriveLaW starts with the most synthetic scenes failing the drivable-area (DAC).
Guided SimWAM is the strongest world action model on the hard-split and stands above every diffusion planner in the table, while DriveVLA-M0 \citep{xing2026drivevlam0} still remains higher.
The drivable-area energy reduces the failures where the plan leaves the road.
The drivable-area score rises in both stages on every model, whereas the sub-metrics that do not concern the road edge move little.
Plans that already stay on the road are left almost unchanged, because the energy only acts when a corner of the vehicle comes within the margin of the road edge, which happens on about $15\%$ of the scenes the models already pass.
The two-frame comfort term (EC) changes by at most $1.8$ points on any model and stage, which is within or close to its standard deviation over seeds.
The calibrated guidance strength $\lambda$ differs by two orders of magnitude between the models: $4$ on DriveWAM against $0.05$ on SimWAM and DriveLaW, which requires a per-model calibration and is analyzed in \S\ref{sec:ablation}.

\begin{table*}[t]
\centering
\scriptsize
\setlength{\tabcolsep}{5pt}
\begin{tabular}{llcrrrrrrrrrr}
\toprule
Method & Venue & Stage & NC$\uparrow$ & DAC$\uparrow$ & DDC$\uparrow$ & TLC$\uparrow$ & EP$\uparrow$ & TTC$\uparrow$ & LK$\uparrow$ & HC$\uparrow$ & EC$\uparrow$ & EPDMS$\uparrow$ \\
\midrule
\multicolumn{13}{l}{\emph{Diffusion and flow planners}} \\
\multirow{2}{*}{TransFuser \citep{chitta2022transfuser}} & \multirow{2}{*}{TPAMI'23} & S1 & 96.2\nosd & 79.5\nosd & 99.1\nosd & 99.5\nosd & 84.1\nosd & 95.1\nosd & 94.2\nosd & 97.5\nosd & 79.1\nosd & \multirow{2}{*}{25.1\nosd} \\
 & & S2 & 77.7\nosd & 70.2\nosd & 84.2\nosd & 98.0\nosd & 85.1\nosd & 75.6\nosd & 45.4\nosd & 95.7\nosd & 75.9\nosd & \\
\multirow{2}{*}{DiffusionDrive \citep{liao2025diffusiondrive}} & \multirow{2}{*}{CVPR'25} & S1 & 96.8\nosd & 86.0\nosd & 98.8\nosd & 99.3\nosd & 84.0\nosd & 95.8\nosd & 96.7\nosd & 97.6\nosd & 79.6\nosd & \multirow{2}{*}{27.5\nosd} \\
 & & S2 & 80.1\nosd & 72.8\nosd & 84.4\nosd & 98.4\nosd & 85.9\nosd & 76.6\nosd & 46.4\nosd & 96.3\nosd & 72.8\nosd & \\
\multirow{2}{*}{GuideFlow \citep{guideflow2025}} & \multirow{2}{*}{CVPR'26} & S1 & 96.6\nosd & 80.5\nosd & 96.3\nosd & 99.3\nosd & 82.3\nosd & 94.9\nosd & 91.5\nosd & 97.7\nosd & 67.8\nosd & \multirow{2}{*}{27.1\nosd} \\
 & & S2 & 87.3\nosd & 76.7\nosd & 88.8\nosd & 99.2\nosd & 84.3\nosd & 85.1\nosd & 49.7\nosd & 93.1\nosd & 44.5\nosd & \\
\midrule
\multicolumn{13}{l}{\emph{Vision-language-action models}} \\
\multirow{2}{*}{ReCogDrive \citep{li2026recogdrive}} & \multirow{2}{*}{ICLR'26} & S1 & 96.4\nosd & 78.9\nosd & 98.7\nosd & 99.8\nosd & 82.6\nosd & 95.6\nosd & 94.4\nosd & 97.6\nosd & 74.2\nosd & \multirow{2}{*}{25.7\nosd} \\
 & & S2 & 80.2\nosd & 65.0\nosd & 82.4\nosd & 98.7\nosd & 85.2\nosd & 76.9\nosd & 43.8\nosd & 96.6\nosd & 71.8\nosd & \\
\multirow{2}{*}{SGDrive \citep{li2026sgdrive}} & \multirow{2}{*}{CVPR'26} & S1 & 95.8\nosd & 87.6\nosd & 97.8\nosd & 99.8\nosd & 84.4\nosd & 94.7\nosd & 92.9\nosd & 97.8\nosd & 28.9\nosd & \multirow{2}{*}{25.5\nosd} \\
 & & S2 & 79.4\nosd & 65.4\nosd & 79.1\nosd & 98.9\nosd & 88.9\nosd & 75.3\nosd & 42.7\nosd & 96.4\nosd & 29.6\nosd & \\
\multirow{2}{*}{DriveFine \citep{dang2026drivefine}} & \multirow{2}{*}{ECCV'26} & S1 & 97.6\nosd & 90.0\nosd & 99.1\nosd & 99.3\nosd & 84.9\nosd & 96.7\nosd & 97.3\nosd & 97.6\nosd & 72.0\nosd & \multirow{2}{*}{30.5\nosd} \\
 & & S2 & 82.1\nosd & 71.3\nosd & 84.8\nosd & 98.4\nosd & 88.1\nosd & 74.3\nosd & 47.2\nosd & 96.8\nosd & 72.8\nosd & \\
\multirow{2}{*}{DriveVLA-M0 \citep{xing2026drivevlam0}} & \multirow{2}{*}{MM'26} & S1 & 98.9\nosd & 96.2\nosd & 99.7\nosd & 100.0\nosd & 73.8\nosd & 99.1\nosd & 94.2\nosd & 97.3\nosd & 64.4\nosd & \multirow{2}{*}{47.0\nosd} \\
 & & S2 & 91.1\nosd & 89.5\nosd & 93.1\nosd & 98.6\nosd & 65.5\nosd & 89.2\nosd & 52.7\nosd & 98.6\nosd & 72.1\nosd & \\
\midrule
\multicolumn{13}{l}{\emph{World action models}} \\
\multirow{2}{*}{DriveLaW \citep{xia2026drivelaw}} & \multirow{2}{*}{CVPR'26} & S1 & 97.1\sd{0.0} & 88.7\sd{0.2} & 99.2\sd{0.1} & 99.6\sd{0.0} & 84.2\sd{0.1} & 97.0\sd{0.1} & 96.2\sd{0.2} & 97.8\sd{0.0} & 63.7\sd{2.7} & \multirow{2}{*}{29.8\sd{0.3}} \\
 & & S2 & 82.2\sd{0.2} & 67.2\sd{0.2} & 83.8\sd{0.2} & 98.1\sd{0.0} & 84.6\sd{0.1} & 78.7\sd{0.0} & 46.9\sd{0.5} & 96.5\sd{0.1} & 57.1\sd{0.7} & \\
\multirow{2}{*}{\quad + EG (ours)} & & S1 & 96.6\sd{0.1} & 94.9\sd{0.2} & 99.1\sd{0.1} & 99.6\sd{0.0} & 84.1\sd{0.1} & 95.7\sd{0.1} & 95.0\sd{0.5} & 97.8\sd{0.0} & 61.9\sd{4.1} & \multirow{2}{*}{33.6\sd{0.4}} \\
 & & S2 & 82.7\sd{0.3} & 71.5\sd{0.4} & 83.8\sd{0.1} & 98.1\sd{0.0} & 84.5\sd{0.1} & 79.7\sd{0.3} & 45.0\sd{0.4} & 96.6\sd{0.0} & 57.8\sd{0.6} & \\
\multirow{2}{*}{DriveWAM \citep{drivewam2026}} & \multirow{2}{*}{arXiv'26} & S1 & 97.2\sd{0.2} & 92.5\sd{0.3} & 99.7\sd{0.1} & 99.3\sd{0.0} & 84.7\sd{0.2} & 95.9\sd{0.1} & 98.4\sd{0.2} & 97.8\sd{0.0} & 80.7\sd{1.0} & \multirow{2}{*}{35.5\sd{0.7}} \\
 & & S2 & 81.9\sd{0.4} & 76.0\sd{0.4} & 85.8\sd{0.5} & 98.3\sd{0.2} & 87.2\sd{0.4} & 77.3\sd{0.3} & 49.8\sd{0.6} & 95.2\sd{0.2} & 72.7\sd{1.2} & \\
\multirow{2}{*}{\quad + EG (ours)} & & S1 & 97.0\sd{0.2} & 95.9\sd{0.1} & 99.5\sd{0.1} & 99.3\sd{0.0} & 84.6\sd{0.2} & 96.6\sd{0.1} & 97.2\sd{0.1} & 97.8\sd{0.0} & 79.9\sd{0.7} & \multirow{2}{*}{37.8\sd{0.8}} \\
 & & S2 & 82.3\sd{0.3} & 77.6\sd{0.5} & 85.3\sd{0.9} & 98.6\sd{0.2} & 87.4\sd{0.3} & 77.5\sd{0.4} & 48.0\sd{0.7} & 95.2\sd{0.2} & 71.4\sd{0.8} & \\
\multirow{2}{*}{SimWAM \citep{simwam2026}} & \multirow{2}{*}{arXiv'26} & S1 & 97.7\sd{0.1} & 91.7\sd{0.8} & 99.6\sd{0.0} & 99.6\sd{0.0} & 83.9\sd{0.1} & 96.2\sd{0.2} & 97.8\sd{0.0} & 97.8\sd{0.0} & 76.6\sd{1.0} & \multirow{2}{*}{37.3\sd{0.6}} \\
 & & S2 & 81.8\sd{0.3} & 78.6\sd{0.1} & 87.5\sd{0.0} & 98.4\sd{0.0} & 86.3\sd{0.1} & 78.5\sd{0.2} & 50.2\sd{0.3} & 96.3\sd{0.1} & 72.1\sd{2.4} & \\
\multirow{2}{*}{\quad + EG (ours)} & & S1 & 97.7\sd{0.1} & 95.7\sd{0.6} & 99.5\sd{0.1} & 99.6\sd{0.0} & 83.9\sd{0.1} & 96.5\sd{0.3} & 98.2\sd{0.4} & 97.8\sd{0.0} & 76.6\sd{1.4} & \multirow{2}{*}{39.6\sd{0.5}} \\
 & & S2 & 82.1\sd{0.3} & 79.8\sd{0.3} & 87.7\sd{0.1} & 98.4\sd{0.0} & 86.5\sd{0.1} & 78.6\sd{0.1} & 50.7\sd{0.1} & 96.2\sd{0.2} & 70.4\sd{1.3} & \\
\bottomrule
\end{tabular}
\caption{NAVSIM v2 hard-split. Per-metric scores of stage one (S1, real scenes) and stage two (S2, synthetic scenes) and the combined EPDMS, in points. \textbf{EG}: energy guidance with a single camera-based drivable-area energy and no privileged information at inference, mean and standard deviation over three seeds. The stock rows of the three WAMs are the mean and standard deviation over three seeds as well. Baseline rows are the numbers published in the respective papers, except TransFuser from the NAVSIM v2 leaderboard, DiffusionDrive as reproduced in SimScale, and ReCogDrive and SGDrive as reproduced in the SimWAM paper. GuideFlow is its no-scorer setting. The stock DriveLaW, DriveWAM and SimWAM rows are our own runs of the released checkpoints.}
\label{tab:main}
\vspace{-8pt}
\end{table*}

\begin{figure*}[t]
\centering
\includegraphics[width=\textwidth]{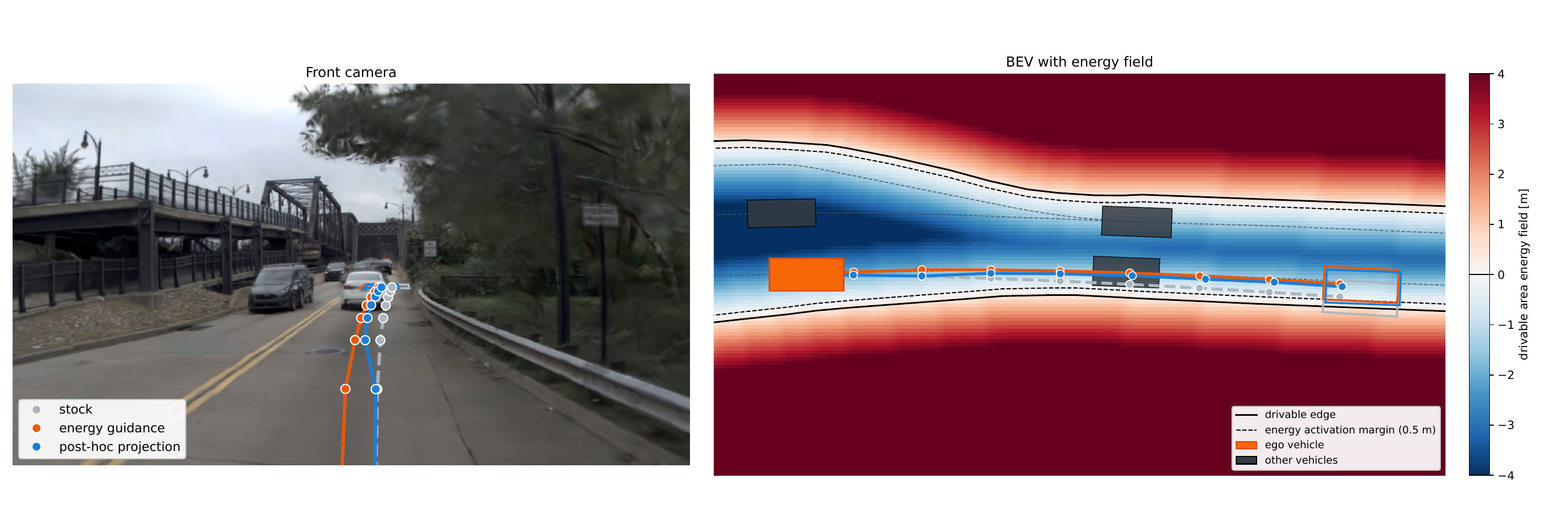}
\caption{Energy guidance against post-hoc projection on one NAVSIM v2 hard-split scene where both recover the DAC. \textbf{Left}: front camera with the three plans projected into the image. \textbf{Right}: bird's-eye view (BEV) with the drivable energy field. The stock plan (grey) leaves the drivable area. The post-hoc projection (blue) follows it until the plan violates the constraint, causing sudden lateral change. Guidance (orange) corrects along the whole plan and reaches the same score with a smooth trajectory.}
\label{fig:case_projection}
\end{figure*}
% \pending{paired bootstrap intervals per model, seed spread on DriveWAM}
\noindent\textbf{In-sampling guidance vs post-hoc projection}
A safety filter applied after generation is the classical way to enforce a constraint.
Table~\ref{tab:projector} applies a gradient projection of the finished stock plan onto the same drivable-area constraint.
As a fix for drivable-area failures, the projector is stronger on all three models: it clears between five and nine points of failing synthetic scenes where guidance clears between one and three, and it scores one to three EPDMS points higher.
However, this is paid in consistency: two-frame comfort (EC) drops by $6.7$ to $7.5$ points under projection, against at most $1.6$ under guidance. 
Adding a smoothness term to the projection objective recovers up to 48\% of the comfort gap and does not reach the comfort of the guided plans.
As guidance moves the plan at every denoising step, the correction is spread along the horizon, and the frozen model keeps shaping the plan after each step, whereas the projector activates on only a third of the scenes and finds no feasible plan on one eighth of them. Its largest corrections are of more than two meters.
On the scenes where both channels recover the drivable-area failure, the guided plans keep a two-frame comfort of $80.7$ against $72.5$ for the projected ones, with the stock plans at $82.6$.
Figure~\ref{fig:case_projection} shows one example of such scenes.
The stock plan drifts toward the right edge of the drivable area and fails.
The projector leaves the plan untouched until the footprint violates the constraint and then bends it back at that waypoint.
Guidance acts on the overall plan, so the guided trajectory leaves the ego vehicle on a smoother heading from the first waypoint, even though the energy only activates once a corner comes within half a meter of the edge.
Both plans correct the failure. The scorer registers the abrupt correction only through the comfort metrics, which the LQR tracker in NAVSIM partly smooths and carry less weight, so the projected plan loses little score for an abrupt change of direction the vehicle would feel.
From the comparison, we can draw two conclusions: Guidance is a viable solution to steer the plan in frozen WAMs, which incorporate new driving objectives with little cost to other metrics, like comfort. However, a post-hoc projection does reach a higher score at the cost of consistency.
% \red{Guidance earns its place for what the filter cannot do: it acts on the action stream while the video is still being generated, which is what makes it the channel that can reach the imagination in \S\ref{sec:controllability}.}
% \pending{rerun the projector on the footprint-only guided arm of DriveWAM and SimWAM, the rows below pair it with the lane-plus-footprint arm of those two models}

\begin{table}[t]
\centering
\scriptsize
\setlength{\tabcolsep}{5pt}
\begin{tabular}{llrrrr}
\toprule
Model &  & EPDMS$\uparrow$ & $\Delta$ & Off-road$\downarrow$ & EC$\uparrow$ \\
\midrule
\multirow{4}{*}{DriveLaW} & stock & $30.0$ & & $26.0$ & $57.9$ \\
 & guidance & $34.8$ & $+4.8$ & $22.7$ & $\mathbf{58.0}$ \\
 & projection & $\mathbf{37.7}$ & $+7.7$ & $\mathbf{16.6}$ & $51.2$ \\
 & \quad + smoothing & $37.3$ & $+7.2$ & $16.7$ & $54.4$ \\
\midrule
\multirow{4}{*}{DriveWAM} & stock & $35.9$ & & $20.9$ & $\mathbf{73.0}$ \\
 & guidance & $39.2$ & $+3.3$ & $18.5$ & $71.4$ \\
 & projection & $\mathbf{40.3}$ & $+4.3$ & $\mathbf{15.2}$ & $66.0$ \\
 & \quad + smoothing & $39.8$ & $+3.8$ & $15.6$ & $66.9$ \\
\midrule
\multirow{4}{*}{SimWAM} & stock & $37.8$ & & $18.7$ & $\mathbf{72.8}$ \\
 & guidance & $39.2$ & $+1.5$ & $18.0$ & $71.6$ \\
 & projection & $\mathbf{41.0}$ & $+3.2$ & $\mathbf{14.0}$ & $65.3$ \\
 & \quad + smoothing & $40.5$ & $+2.8$ & $\mathbf{14.0}$ & $67.0$ \\
\bottomrule
\end{tabular}
\caption{Guidance vs post-hoc gradient projection with the same constraint. Both use the privileged drivable area with a margin of $0.5$\,m. Off-road is the fraction of synthetic scenes (stage 2) failing the DAC. EC is the stage-two two-frame comfort.}
\label{tab:projector}
\end{table}

\noindent\textbf{Temporary Speed limits}
To test an objective beyond the drivable area, we guide the same frozen models with a speed cap, simulating a temporary speed limit.
Table~\ref{tab:speedcap} shows a cap of $5$\,m/s, i.e. $18$km/h.
The percentage of plans that exceed the cap falls from about $48\%$ to $0.2\%$ on DriveWAM, $2.1\%$ on DriveLaW and $9.9\%$ on SimWAM.
Ego progress falls with it, since the benchmark rewards progress, and hence so does the EPDMS.
Combined with the drivable-area energy, the cap keeps most of its effect and the drivable-area compliance rises further.

\begin{table}[t]
\centering
\scriptsize
\setlength{\tabcolsep}{4pt}
\begin{tabular}{llrrrr}
\toprule
Model & & Over limits$\downarrow$ & EP$\uparrow$ & DAC$\uparrow$ & EPDMS$\uparrow$ \\
\midrule
\multirow{3}{*}{DriveLaW} & stock & $47.3$ & $84.7$ & $67.4$ & $30.0$ \\
 & speed cap & $\mathbf{2.1}$ & $74.2$ & $75.3$ & $28.3$ \\
 & speed cap + DA & $6.5$ & $74.4$ & $\mathbf{78.5}$ & $30.6$ \\
\midrule
\multirow{3}{*}{DriveWAM} & stock & $49.2$ & $87.0$ & $76.4$ & $35.9$ \\
 & speed cap & $\mathbf{0.2}$ & $73.8$ & $80.8$ & $32.1$ \\
 & speed cap + DA & $0.5$ & $73.8$ & $\mathbf{83.8}$ & $35.5$ \\
\midrule
\multirow{3}{*}{SimWAM} & stock & $47.2$ & $86.2$ & $78.5$ & $37.8$ \\
 & speed cap & $\mathbf{9.9}$ & $76.8$ & $84.0$ & $33.7$ \\
 & speed cap + DA & $11.0$ & $76.9$ & $\mathbf{84.2}$ & $34.8$ \\
\bottomrule
\end{tabular}
\caption{Speed cap of $5$\,m/s ($18$\,km/h) as a sampling-time objective. Over limits: the percentage of plans with at least one waypoint speed above the cap.}
\label{tab:speedcap}
\end{table}

\subsection{Ablation}
\label{sec:ablation}

\begin{figure}[t]
\centering
\includegraphics[width=\linewidth]{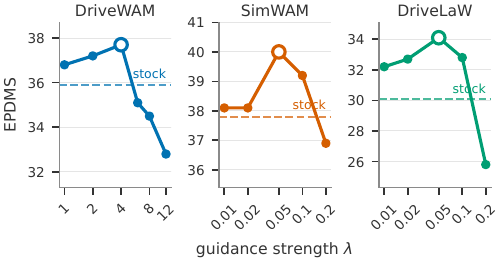}
\caption{Guidance strength on single-camera-only drivable-area energy. EPDMS against the strength $\lambda$ on the logarithmic x-axis. The dashed lines show the stock models. The open marker is the used strength in Table~\ref{tab:main}.}
\label{fig:strength}
\end{figure}

\begin{figure}[t]
\centering
\includegraphics[width=0.8\linewidth]{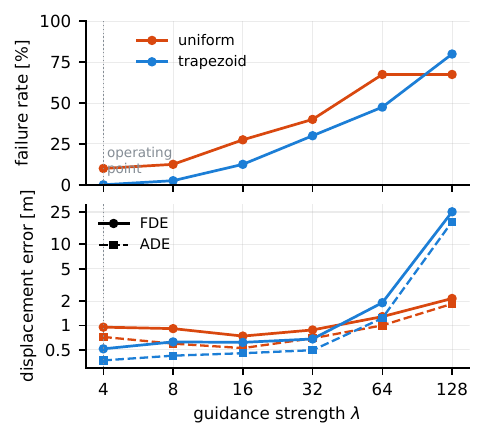}
\caption{Window shape against guidance strength on $40$ DriveWAM DAC failure scenes under the stock sampler. \textbf{Top}: failure rate, the percentage of scenes whose final plan jitters more than three times as the stock sampler. \textbf{Bottom}: ADE/FDE against the stock plan.}
\label{fig:window}
\end{figure}

\textbf{Energy composition}
Energies can be composed as the weighted sum of Eq.~\eqref{eq:norm}. As NAVSIM allows limited information for energy construction, Table~\ref{tab:main} uses only one legal energy: camera-based drivable area (DA). To further investigate the energy composition, Table~\ref{tab:combo} uses privileged information to construct two additional energy terms: driving direction (DD) and lane keeping (LK).
The DD energy penalizes the distance driven outside the lanes of the route.
Alone, it is nearly inert, within one point of stock on every model. On top of the DA energy, it adds $0.2$ to $0.8$ points on each model. 
DA + DD scores best on DriveLaW and SimWAM. On DriveWAM, the three compositions with the DA energy lie within $0.2$ points of each other.
The sub-metrics show where the gain comes from. The DDC of the stage 2 in NAVSIM rises by $1$ to $4$ points, but it costs the extended comfort term, which drops by $4$ to $11$.
\begin{table}[t]
\centering
\scriptsize
\setlength{\tabcolsep}{3pt}
\begin{tabular}{lrrrrrrr}
\toprule
 & & \multicolumn{3}{c}{single energy EPDMS$\uparrow$} & \multicolumn{3}{c}{composition EPDMS$\uparrow$} \\
\cmidrule(lr){3-5}\cmidrule(lr){6-8}
Model & stock & LK & DA & DD & LK+DA & DA+DD & LK+DA+DD \\
\midrule
DriveLaW & $30.0$ & $31.1$ & $34.8$ & $30.9$ & $33.3$ & $\mathbf{35.6}$ & $33.0$ \\
DriveWAM & $35.9$ & $37.5$ & $39.2$ & $36.3$ & $39.5$ & $39.4$ & $\mathbf{39.6}$ \\
SimWAM   & $37.8$ & $37.9$ & $39.2$ & $38.4$ & $38.2$ & $\mathbf{39.9}$ & $38.9$ \\
\bottomrule
\end{tabular}
\caption{Composition of guidance energies on the privileged information. Lane keeping (LK): centerline attraction; drivable area (DA): drivable-area violation on the vehicle corners; driving direction (DD): distance driven outside the lanes of the route.}
\label{tab:combo}
\end{table}
Interestingly, omposing all three energies is not better than DA + DD on DriveLaW and SimWAM, as the DA and DD are sparse while LK is dense and continuous, which pulls on every scene, costing comfort on passed scenes. \\
\textbf{Guidance strength on planning policy}
Figure~\ref{fig:strength} sweeps the guidance strength on the DA energy over 5 values per model.
DriveWAM improves on stock between $\lambda$ of $1$ and $4$ and falls below it at $6$, SimWAM and DriveLaW improve between $0.01$ and $0.10$ and collapse at $0.20$, where DriveLaW loses four points against stock.
The strengths differ by two orders of magnitude.
SimWAM and DriveLaW peak at the same relative correction, about $1.4\%$ of the model velocity, at a strength of $0.05$.
DriveWAM peaks at $45\%$ at a strength of $4$, thirty times more. Its relative correction grows from $13\%$ at a strength of $1$ to $91\%$ at $12$.
\begin{table}[t]
\centering
\scriptsize
\setlength{\tabcolsep}{8pt}
\begin{tabular}{lrrr}
\toprule
Model & full-split best & cross-fitted & 95\% interval \\
\midrule
DriveLaW & $34.1$ & $31.4$ & $[+1.9, +6.3]$ \\
DriveWAM & $38.2$ & $37.7$ & $[+0.5, +4.2]$ \\
SimWAM & $40.0$ & $40.0$ & $[+0.8, +3.8]$ \\
\bottomrule
\end{tabular}
\caption{Cross-fitted selection of strength and margin over two folds of the $225$ scene groups, next to the full-split optimum. The $95$\% interval is calculated from $10{,}000$ random redraws of $450$ scenes.}
\label{tab:crossfit}
\end{table}
As NAVSIM v2 hard-split has no validation split, to investigate the method's generalization, we therefore conduct a cross-fitted search study by splitting the $450$ real observations from stage 1 into two folds. The guidance strength $\lambda$ as well as energy hyperparameters are searched on one fold, where the EPDMS is evaluated on the other. We swap the folds and average them. 
Table~\ref{tab:crossfit} gives the result compared with the full-split optimum.
The cross-fitted gains are $1.7$, $2.2$ and $1.4$ points on DriveWAM, SimWAM and DriveLaW. \\
\textbf{Guidance to noise level}
Figure~\ref{fig:window} measures the influence on $40$ DriveWAM failure scenes on drivable area compliance (DAC) under its stock sampler.
We measure the off-manifold failure as the final denoised plan jitters more than three times the stock sampler does, which signals a sampler being pushed off its manifold. We measure the average displacement error (ADE) and final displacement error (FDE) of the guided plan compared with the stock plan \citep{alahi2016social}. 
With a uniform weight, some scenes already jitter at a strength of $4$, and the percentage rises with the strength (Table~\ref{tab:stab}).
With the trapezoid, no scene jitters at $4$ and the fraction stays below the uniform schedule up to $32$.
The ADE/FDE show similar trends. Within the usable range, the trapezoid moves the plan endpoint by about half a meter, whereas the uniform weight moves it by nearly one, so the scheduled correction is both smaller and more stable.
At $128$, the trapezoid is worse, because its weight in the late window is still large when the plan is settled.
The real-scene stage (stage 1) falls from $84$ to $71$ and the DDC drops five points, which is the off-manifold indicator. \\
\textbf{Stabilization components}
\begin{table}[t]
\centering
\scriptsize
\setlength{\tabcolsep}{4pt}
\begin{tabular}{lrrrr}
\toprule
 & full-split & \multicolumn{3}{c}{failure rate [\%] on $40$ scenes} \\
\cmidrule(lr){3-5}
Variant & EPDMS$\uparrow$ & $\lambda = 4$ & $\lambda = 16$ & $\lambda = 64$ \\
\midrule
full method & $\mathbf{37.7}$ & $\mathbf{0.0}$ & $\mathbf{7.5}$ & $42.5$ \\
w.o. guidance scheduling & $31.2$ & $5.0$ & $10.0$ & $\mathbf{37.5}$ \\
w.o. velocity bound & $\mathbf{37.7}$ & $\mathbf{0.0}$ & $\mathbf{7.5}$ & $60.0$ \\
\bottomrule
\end{tabular}
\caption{Ablation study for stabilization mechanisms on full hard-split and $40$ DriveWAM failure scenes. Full-split: strength $\lambda = 4$. Failure rate: percentage of the $40$ drivable-area failure scenes. The stock model scores $35.9$.}
\label{tab:stab}
\end{table}
To ablate the stabilization mechanisms, we conduct the study on the same 40 DriveWAM failure scenes, shown in Table~\ref{tab:stab}. On the full hard-split of DriveWAM, the uniform window scores $31.2$ against $37.7$ for the trapezoid.
The velocity bound is not active at the used guidance strength, but reduces failure rate at high guidance strength.
At a strength of $64$, removing it raises the failure rate on the $40$ scenes from $42.5\%$ to $60\%$.
The window therefore carries the stability at the operating points, whereas the velocity bound protects against divergence above them.\\
\textbf{Jacobian-free backpropagation}
To investigate the influence of eliminating the network Jacobian in Eq.~\ref{eq:jacobian}, we compare the Jacobian and the Jacobian-free forms for three models and measure the EPDMS, latency, and memory usage in Table~\ref{tab:jacobian}. 
On SimWAM, the action expert is small enough for a backward pass through the frozen network at every sampler step.
At the same guidance strength, the Jacobian form scores lower on all three models.
The difference shows in the latency: the Jacobian form spends $14$ milliseconds per step on guidance against $1.4$ for the Jacobian-free form, which equals a full extra backward pass on top of the $16$ millisecond forward pass, at the same peak memory.
On DriveWAM, the Jacobian-free correction adds $1.9$ milliseconds on a $36$ millisecond forward pass. The backpropagation costs $72$ milliseconds through the five-billion-parameter transformer.
Guidance without backpropagation therefore adds under $10\%$ to the sampling time of either model, whereas the Jacobian form adds more than $88\%$ of a forward pass.

\begin{table}[t]
\centering
\scriptsize
\setlength{\tabcolsep}{3pt}
\begin{tabular}{llrrrr}
\toprule
Model & Form & EPDMS$\uparrow$ & guidance / step$\downarrow$ & forward / step$\downarrow$ & memory$\downarrow$ \\
\midrule
DriveLaW & Jacobian-free & $34.1$ & $0.1$\,ms & $48$\,ms & $8.2$\,GB \\
 & Jacobian & $29.9$ & $50.4$\,ms & $42.4$\,ms & $9.8$\,GB \\
DriveWAM & Jacobian-free & $37.7$ & $1.9$\,ms & $36$\,ms & $13.7$\,GB \\
 & Jacobian & $36.3$ & $72.5$\,ms & $35.9$\,ms & $14.3$\,GB \\
SimWAM & Jacobian-free & $40.0$ & $1.4$\,ms & $16.3$\,ms & $12.7$\,GB \\
 & Jacobian & $37.9$ & $14.3$\,ms & $16.2$\,ms & $12.7$\,GB \\
\bottomrule
\end{tabular}
\caption{Jacobian-free guidance against the full Jacobian form of Eq.~\eqref{eq:jacobian}, at the same strength and window. Times are per sampler step on one NVIDIA RTX 4090 GPU. Memory is the peak during sampling.}
\label{tab:jacobian}
\end{table}

%% file: sections/05_conclusion.tex
\section{Limitations}
\label{sec:limitations}

Our evaluation is open-loop pseudo-simulation on one benchmark, with no closed-loop test. This is due to the sim-to-real gap of existing closed-loop frameworks such as Bench2Drive \citep{jia2024bench}.
Since the guidance is soft, the energy terms can only reduce violations but not guarantee that a driving objective is enforced.
Composition of different energies is done by a weighted sum, which cannot reconcile two contradicting energy terms. A rule-aware combination such as the signal temporal logic robustness in CTG \citep{zhong2023ctg} is the natural solution.
All three models are sampled with a deterministic Euler integrator, so the schedule and the operating range are untested under stochastic samplers in case of a stochastic backbone.

\section{Conclusion}
\label{sec:conclusion}

We show that frozen world action models can be guided by energies on their one-step clean estimate at sampling time to follow new driving objectives.
In practice, the guidance can be stabilized empirically by a trapezoidal window over the noise level, with a correction bounded relative to the model's own velocity as a safeguard.
We show performance improvement on three WAMs: DriveLaW, DriveWAM, and SimWAM on NAVSIM v2 hard-split over stock samplers. 
Compared with a post-hoc projection, the guidance mechanism trades only a few points in EPDMS for better comfort in NAVSIM v2 hard-split.
The studies on energy composition, guidance strength, and computation overhead further identify the energy guidance as a practical post-training control channel for WAMs.

%% file: sections/99_acknowledgement.tex
% \section*{AI use statement}

% Large language models were used as coding and writing assistants throughout this project.
% All experimental designs, claims, and final text were reviewed and are the responsibility of the authors.

% \section*{Reproducibility}
% The code for all experiments, the configuration and seeds of every table, the scene lists of the ablations, and the checkpoint of the road-edge network are available at \url{https://github.com/TUMFTM/GuidedWAM} (version v1.0).
% The three world action models are the checkpoints released by their authors, and the repository states how to obtain them.

% \section*{Acknowledgments}

%% file: sections/A_supplementary.tex
% Appendix of the arXiv version. Same content as the supplementary material of the workshop paper.
\clearpage
\appendix

\noindent The appendix describes the form of the energies, the settings of the guided sampler, the camera-only road-edge network, and the projection baseline.
It also provides additional results and further qualitative results.
All definitions follow the released code on \url{https://tumftm.github.io/GuidedWAM/}.

\section{Notation}
\label{sec:supp_notation}

A plan is the trajectory of $T = 8$ poses at $\Delta t = 0.5$\,s in the current ego frame, with the rear axle as origin, $x$-axis pointing forward and $y$-axis pointing to the left,
\begin{equation}
\begin{aligned}
\tau &= (p_1, \dots, p_T) \\
p_t &= (x_t, y_t, \psi_t) \\
p_0 &= (0, 0, 0)
\end{aligned}
\end{equation}
Every energy maps a plan to a non-negative scalar and is differentiable in $\tau$.
We write $\pos{u} = \max(u, 0)$ and use the gradient $0$ at $u = 0$.

\textbf{Interpolated path}
The drivable-area and driving-direction energies are evaluated on an interpolated path, which approximates the one the scorer of NAVSIM~v2 simulates between the waypoints
The current pose $p_0$ is prepended and $(x, y, \psi)$ is interpolated linearly with four points per segment.
This gives $M + 1 = 33$ poses $q_0, \dots, q_M$ with $M = 32$, from the current pose $q_0 = p_0$ to the last waypoint $q_M = p_T$.

\textbf{Ego footprint}
Following the vehicle parameters of nuPlan~\citep{caesar2021nuplan}, which the scorer also uses, the vehicle has length $L = 5.176$\,m, width $W = 2.297$\,m and a distance of $a = 1.461$\,m from the rear axle to the centre of its box.
Its four corners relative to the rear axle are
\begin{equation}
\mathcal{C} = \big\{(a + \tfrac{L}{2}, \pm\tfrac{W}{2}),\ (a - \tfrac{L}{2}, \pm\tfrac{W}{2})\big\}
\end{equation}
At a pose $q_m = (x_m, y_m, \psi_m)$, a corner $c \in \mathcal{C}$ lies at $r_{m,c} = (x_m, y_m) + R(\psi_m)\, c$, with $R$ the planar rotation.

\textbf{Signed-distance fields}
$\phi(r)$ is the signed distance in meters from a point $r$ to the boundary of a region.
It is negative inside and positive outside.

\section{Energies}
\label{sec:supp_energies}

\begin{figure*}[t]
\centering
\includegraphics[width=0.9\textwidth]{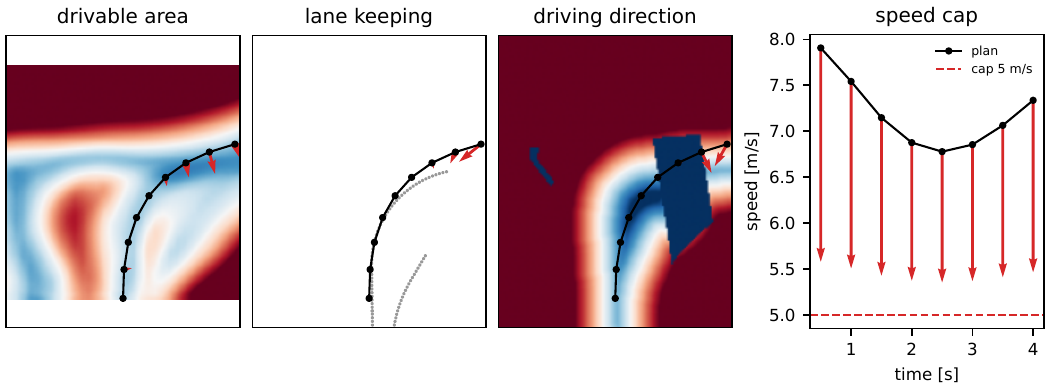}
\caption{The four energies on one plan. The first three panels show the field or the lane points that the energy reads, the plan, and the direction in which the gradient moves the waypoints (red). In the two field panels, blue is inside the region and red is outside. In the driving-direction panel, the blue band is the lanes of the route, where the energy is zero, and red is the area outside them, including the oncoming lanes. The uniformly dark blue areas are intersections, which the scorer excludes from the driving-direction compliance (DDC) and where the energy is therefore switched off. The last panel shows the planned speed against a cap of $5$\,m/s.}
\label{fig:supp_energies}
\end{figure*}

\subsection{Drivable area (DA)}
The scorer fails the drivable-area compliance of a plan as soon as one footprint corner leaves the drivable area at one simulated step.
The energy penalizes this quantity with a margin $\delta$,
\begin{equation}
E_{\mathrm{DA}}(\tau) = \frac{w}{4(M+1)} \sum_{m=0}^{M} \sum_{c \in \mathcal{C}} \pos{\phi_{\mathrm{DA}}(r_{m,c}) + \delta}
\label{eq:supp_da}
\end{equation}
A corner more than $\delta$ inside the edge contributes nothing, and the penalty grows linearly as the corner approaches or crosses the edge.
The energy depends on the heading through $R(\psi_m)$, so its gradient also turns the vehicle.
The field $\phi_{\mathrm{DA}}$ comes from one of two sources.

\textbf{Privileged field}
The region is the union of the map layers the scorer uses for the drivable area: roadblocks, roadblock connectors, intersections and car parks.
It is rasterized for sampling speed and computed on $x \in [-10, 60)$\,m and $y \in [-25, 25)$\,m for plan coverage.
The margin is $\delta = 0.5$\,m.
This field uses the nuPlan map, which is considered as privileged information by NAVSIM, and is therefore only used for the comparison with projection, the composition study and the window study  \citep{caesar2021nuplan,cao2025navsimv2}.

\textbf{Learned field}
The field is the output of the road-edge network of Section~\ref{sec:supp_field} with its own grid of $x \in [0, 32]$\,m and $y \in [-32, 32]$\,m.
Corners outside this grid are unknown and receive no penalty and no gradient.
The margin is $\delta = 1.0$\,m for DriveWAM and $1.5$\,m for SimWAM and DriveLaW.
The detected road-edge tends to be smaller than the true one. Therefore, we use a higher margin to compensate the detected violations (Table~\ref{tab:supp_corner}).
This field is used for the main results, the strength study, and the cross-fitted study.

\subsection{Lane keeping (LK)}
Each waypoint is attracted to the nearest lane centerline point, with a dead zone of $d_0 = 0.5$\,m.
With $\mathbf{u}_t = (x_t, y_t)$ and centerline points $\ell_1, \dots, \ell_J$ in the ego frame,
\begin{equation}
\begin{aligned}
E_{\mathrm{LK}}(\tau) &= \frac{w}{T} \sum_{t=1}^{T} \pos{d_t - d_0}^2 \\
d_t &= \min_{j} \lVert \mathbf{u}_t - \ell_j \rVert_2
\end{aligned}
\end{equation}
The centerline points are sampled every $0.5$\,m along the lanes and lane connectors within $60$\,m of the ego.
Only points with $x \in (-10, 70)$\,m, $y \in (-30, 30)$\,m and a lane heading within $80^\circ$ of the ego heading are kept.
The heading filter keeps oncoming lanes out.
The energy is considered as privileged information as it uses map information.

\subsection{Driving direction (DD)}
The DD energy penalizes driving against the direction of the route. 
Following the scorer, it measures this as the distance the vehicle center travels outside the lanes of the route, which excludes the oncoming lanes.
The energy follows this measure and weights each piece of the path by how far outside the lanes it lies.
With the box centre $o_m = (x_m + a\cos\psi_m,\ y_m + a\sin\psi_m)$ at each pose of the interpolated path,
\begin{equation}
E_{\mathrm{DD}}(\tau) = w \sum_{m=1}^{M} \underbrace{\lVert o_m - o_{m-1} \rVert_2}_{\text{length of the piece}}\ \underbrace{\pos{\phi_{\mathrm{DD}}(o_m)}}_{\text{distance outside}}
\end{equation}
The energy is zero while the center stays in the lanes of the route.
$\phi_{\mathrm{DD}}$ is the signed distance to the lanes of the route, built like the privileged drivable-area field.
Inside intersections it is set to $-8$\,m, since the scorer does not count them.
The energy is considered as privileged information as it uses map and route information.

\subsection{Speed cap}
The speed of the plan is the finite difference between consecutive waypoints,
\begin{equation}
v_t = \frac{\lVert (x_t, y_t) - (x_{t-1}, y_{t-1}) \rVert_2}{\Delta t} \qquad t = 1, \dots, T
\end{equation}
The energy is a squared hinge above a cap $v_{\max}$ that is given at inference,
\begin{equation}
E_{v}(\tau) = w \sum_{t=1}^{T} \pos{v_t - v_{\max}}^2
\end{equation}
The cap is a single number, so the energy does not need map or perception input.

\textbf{Preconditioning}
SimWAM and DriveLaW denoise absolute poses.
There, the gradient of $E_v$ couples neighbouring waypoints, and a plain step mostly moves the last waypoint and makes the plan oscillate.
For these two models, the gradient $g = \partial E_v / \partial \tau$ is preconditioned before it is mapped to the latent space,
\begin{equation}
\tilde g_t = \sum_{s=1}^{t} \sum_{r=s}^{T} g_r
\end{equation}
which is the position change of a descent step in the displacements between waypoints.
DriveWAM denoises pose differences and needs no preconditioning.

% \subsection{Composition}
% Composed arms add weighted energies under one strength $\lambda$.
% Table~\ref{tab:supp_weights} lists the weights.
% $E_{\mathrm{DD}}$ is a sum over the path length and about $100$ times larger than $E_{\mathrm{DA}}$, hence its small weights.

% \begin{table}[t]
% \centering
% \small
% \setlength{\tabcolsep}{4pt}
% \begin{tabular}{lccc}
% \toprule
% Arm & $w_{\mathrm{LK}}$ & $w_{\mathrm{DA}}$ & $w_{\mathrm{DD}}$ \\
% \midrule
% LK & $1$ & & \\
% DA & & $1$ & \\
% DD & & & $1$ \\
% LK + DA & $1$ & $1$ ($2$) & \\
% DA + DD & & $1$ & $0.06$ \\
% LK + DA + DD & $1$ & $1$ ($2$) & $0.12$ ($0.15$) \\
% \bottomrule
% \end{tabular}
% \caption{Relative weights of the composed energies. Values in parentheses: $w_{\mathrm{DA}} = 2$ on DriveLaW, $w_{\mathrm{DD}} = 0.15$ on SimWAM and DriveLaW.}
% \label{tab:supp_weights}
% \end{table}

\section{Guided sampler}
\label{sec:supp_sampler}

All three models are flow-matching samplers with $x_\sigma = (1-\sigma)\, x_0 + \sigma\, \varepsilon$ and velocity target $v = \varepsilon - x_0$, integrated by Euler steps from $\sigma = 1$ to $0$.
Algorithm~\ref{alg:supp_sampler} gives the guided sampler.
The operator $\mathrm{sg}$ stops the gradient, so nothing is backpropagated through the network.
The map $f$ takes the action latents of the model to the metric plan $\tau$.
For DriveWAM, $f$ undoes the normalization of the pose differences and takes a cumulative sum.
The velocity bound in line~6 is applied per sample with $c = 12$.

\begin{algorithm}[t]
\caption{Guided sampling of a frozen WAM}
\label{alg:supp_sampler}
\begin{algorithmic}[1]
\Require frozen velocity model $v_\theta$, context $X$, energy $E$, window $\lambda(\sigma)$, bound $c$, noise levels $1 = \sigma_0 > \dots > \sigma_K = 0$
\State $x^{(0)} \sim \mathcal{N}(0, I)$
\For{$k = 0, \dots, K-1$}
  \State $\hat v_k \gets v_\theta(x^{(k)}, \sigma_k, X)$ \Comment{one forward pass}
  \State $\hat x_0 \gets x^{(k)} - \sigma_k\, \mathrm{sg}(\hat v_k)$ \Comment{clean estimate}
  \State $g \gets \nabla_{\hat x_0} E\big(f(\hat x_0)\big)$ \Comment{energy gradient}
  \State $g \gets g \cdot \min\!\Big(1,\ \dfrac{c\, \lVert \hat v_k \rVert_2}{\lambda(\sigma_k)\, \lVert g \rVert_2}\Big)$ \Comment{velocity bound}
  \State $\tilde v_k \gets \hat v_k + \lambda(\sigma_k)\, g$ \Comment{guided velocity}
  \State $x^{(k+1)} \gets x^{(k)} + \tilde v_k\, (\sigma_{k+1} - \sigma_k)$ \Comment{Euler step}
\EndFor
\State \Return plan $\tau = f(x^{(K)})$
\end{algorithmic}
\end{algorithm}

\textbf{Window}
The strength is a trapezoid over the noise level,
\begin{equation}
\lambda(\sigma) = \lambda \cdot
\begin{cases}
0 & \sigma \ge 0.95 \ \text{or}\ \sigma \le 0.05, \\
(0.95 - \sigma)/0.15 & 0.8 < \sigma < 0.95, \\
1 & 0.3 \le \sigma \le 0.8, \\
(\sigma - 0.05)/0.25 & 0.05 < \sigma < 0.3 .
\end{cases}
\end{equation}

\textbf{Settings}
Table~\ref{tab:supp_settings} lists the operating points of the main results.

\begin{table}[t]
\centering
\small
\setlength{\tabcolsep}{5pt}
\begin{tabular}{lcccc}
\toprule
Model & $\lambda$ & $\delta$ & Steps \\
\midrule
DriveWAM & $4$ & $1.0$\,m & $10$ \\
SimWAM & $0.05$ & $1.5$\,m & $20$ \\
DriveLaW & $0.05$ & $1.5$\,m & $10$  \\
\bottomrule
\end{tabular}
\caption{Sampling and guidance parameters with the learned drivable-area field. DriveWAM runs three video steps before its ten action steps.}
\label{tab:supp_settings}
\end{table}

\section{Camera-only road-edge network}
\label{sec:supp_field}

The network predicts $\phi_{\mathrm{DA}}$ from the cameras of the current frame.
It sees no map, no LiDAR and no route at inference.
The world action models stay frozen.
Only this network is trained.

\subsection{Architecture}
The base is the Latent TransFuser~\citep{chitta2022transfuser} of NAVSIM~\citep{dauner2024navsim}: a ResNet-34 image encoder, a learned latent bird's-eye-view input in place of the LiDAR branch, transformer fusion blocks and a bird's-eye-view decoder.
Only the last $1 \times 1$ convolution of the decoder changes, from seven classes to one regression channel.
The output is a map of $128 \times 256$ cells at $0.25$\,m.
We train from the released checkpoint of the Latent TransFuser.

\textbf{Inputs}
The left, front and right cameras of the current frame are cropped, stitched side by side and resized to $1024 \times 256$ pixels.
A status vector of eight values containing the driving command, the ego velocity and the ego acceleration is appended.

\subsection{Targets}
For every training scene, the polygons of roadblocks, roadblock connectors, intersections and car parks around the ego are rasterized on the output grid.
This gives a mask that marks every cell as drivable or not.
The target of a cell is its signed distance to the road edge, $\phi = d_{\mathrm{out}} - d_{\mathrm{in}}$, where $d_{\mathrm{out}}$ is the distance to the nearest drivable cell and $d_{\mathrm{in}}$ the distance to the nearest non-drivable cell.
One of the two is zero in every cell, so $\phi$ is negative inside the drivable area and positive outside. 

\subsection{Loss and training}
The network is trained to regress the signed distance in every cell.
With $e_{ij} = \hat\phi_{ij} - \phi_{ij}$ the error in cell $(i, j)$ of the output grid $\Omega$, the loss is
\begin{equation}
\begin{aligned}
\mathcal{L} &= \frac{1}{|\Omega|} \sum_{(i,j) \in \Omega} w_{ij}\, \ell(e_{ij}) \\
w_{ij} &= 1 + 4 \exp(-|\phi_{ij}|)
\end{aligned}
\end{equation}
where $\ell$ is the smooth $L_1$ loss, which is quadratic for $|e| < 0.5$\,m and linear above.
The weight $w_{ij}$ falls from $5$ at the road edge to $1$ far from it, since the energy reads the field only near the edge.
The training data is the navtrain split of NAVSIM~\citep{dauner2024navsim}, $103{,}288$ frames from $1{,}192$ logs.
Five percent of the logs are held out for validation, split by log.
We train for $20$ epochs with AdamW, a learning rate of $10^{-4}$, a weight decay of $10^{-4}$, a one-cycle schedule with $5\%$ warm-up, a batch size of $32$ and gradient clipping at $1.0$.
One epoch takes about $12$ minutes on one RTX 4090.
Table~\ref{tab:supp_val} gives the validation results.

\begin{table}[t]
\centering
\small
\begin{tabular}{lr}
\toprule
Validation metric & Value \\
\midrule
MAE, all cells & $0.32$\,m \\
MAE within $1$\,m of the edge & $0.40$\,m \\
Sign accuracy & $97.6\%$ \\
IoU of the drivable area & $94.5\%$ \\
\bottomrule
\end{tabular}
\caption{Road-edge network on held-out navtrain logs.}
\label{tab:supp_val}
\end{table}

\subsection{Inference and cost}
The field is computed once per scene from the current frame, before sampling.
The sampler reads it by bilinear interpolation at every guided step, and the network is not run again.
The network has $56.0$\,M parameters.
On one RTX 4090, one forward pass takes $10.4$\,ms, or $12.0$\,ms with the image preprocessing, at a peak memory of $0.36$\,GB.

\subsection{Quality on the evaluation split}
Table~\ref{tab:supp_quality} reports the field on $1{,}199$ scenes of the hard split where the stock DriveWAM plan fails the drivable-area compliance.
The field is more accurate on real images than on the rendered images of stage two, which it never saw in training.
Table~\ref{tab:supp_corner} counts how many footprint corners that violate the drivable area of the scorer are seen by the energy, and how many compliant corners it flags.

\begin{table}[t]
\centering
\small
\setlength{\tabcolsep}{3pt}
\begin{tabular}{lrrrr}
\toprule
Subset & MAE & MAE $<1$\,m & Sign & IoU \\
\midrule
All & $0.91$\,m & $1.20$\,m & $91.9\%$ & $76.1\%$ \\
Stage one (real) & $0.45$\,m & $0.72$\,m & $96.9\%$ & $89.8\%$ \\
Stage two (rendered) & $0.92$\,m & $1.20$\,m & $91.8\%$ & $75.9\%$ \\
\bottomrule
\end{tabular}
\caption{Road-edge network on drivable-area failure scenes of the hard split. MAE $<1$\,m is the error within one meter of the edge.}
\label{tab:supp_quality}
\end{table}

\begin{table}[t]
\centering
\small
\setlength{\tabcolsep}{3pt}
\begin{tabular}{lrrr}
\toprule
Field & $\delta$ & Detected & False alarms \\
\midrule
Privileged & $0$ & $100.0\%$ & $0.0\%$ \\
Original decoder & -- & $16.9\%$ & $1.3\%$ \\
Learned & $0.0$\,m & $23.0\%$ & $1.2\%$ \\
Learned & $0.5$\,m & $42.2\%$ & $5.1\%$ \\
Learned & $1.0$\,m & $62.2\%$ & $12.2\%$ \\
Learned & $1.5$\,m & $75.4\%$ & $22.0\%$ \\
\bottomrule
\end{tabular}
\caption{Percentage of violating footprint corners detected, and percentage of compliant corners flagged, over $16{,}407$ violating and $70{,}677$ compliant corners.}
\label{tab:supp_corner}
\end{table}

\section{Projection baseline}
\label{sec:supp_proj}

The projection baseline acts on the finished stock plan $\tau_0$ and uses the same footprint-corner constraint as the drivable-area energy on the privileged field, with $\delta = 0.5$\,m.
It minimizes, over a correction $\Delta$ on position and heading,
\begin{equation}
\mu\, E_{\mathrm{DA}}(\tau_0 + \Delta) + \lVert \Delta \rVert_2^2 + \eta\, S(\tau_0 + \Delta)
\end{equation}
with Adam at a learning rate of $0.01$ for at most $1{,}000$ iterations.
The penalty weight $\mu$ escalates through $1$, $10$, $100$ and $1{,}000$ whenever the violation stops decreasing.
The optimization stops when every corner satisfies $\phi_{\mathrm{DA}} + \delta \le 0$.
Plans without a violation pass through unchanged, and plans that still violate at the last iteration keep the stock trajectory.
$S$ is the sum of squared second differences of the waypoints.
The projector of Table~\ref{tab:projector} uses $\eta = 0$, and its smoothed variant uses $\eta = 1$.

\section{Additional results}
\label{sec:supp_results}

All numbers are on the full hard split with the official scorer unless stated otherwise.
Sub-metrics are stage-two values in points.

\subsection{Speed cap}
\label{sec:supp_speed}
Table~\ref{tab:speedcap} reports a cap of $5$\,m/s. Table~\ref{tab:supp_speed} adds a cap of $7.5$\,m/s on all three models, alone and together with the camera-only drivable-area energy.
A plan violates the cap if one of its waypoint speeds exceeds the cap by more than $0.1$\,m/s.
The strength is selected on $300$ random scenes of the split, and the weight of the speed cap in the combined arm keeps this strength.
The two-frame comfort (EC) falls with the cap, because the plan has to decelerate within its horizon.
At the cap of $5$\,m/s it falls from $73.0$ to $48.6$ on DriveWAM, from $72.8$ to $56.2$ on SimWAM and from $57.9$ to $40.5$ on DriveLaW.
At $7.5$\,m/s the loss is smaller, as is the loss in ego progress (EP).

\begin{table}[t]
\centering
\scriptsize
\setlength{\tabcolsep}{3pt}
\begin{tabular}{llrrrrr}
\toprule
Model & Arm & Viol.$\downarrow$ & EP$\uparrow$ & EC$\uparrow$ & DAC$\uparrow$ & EPDMS$\uparrow$ \\
\midrule
\multirow{3}{*}{DriveWAM} & stock & $22.7$ & $87.0$ & $73.0$ & $76.4$ & $35.9$ \\
 & cap & $0.0$ & $82.6$ & $63.1$ & $78.0$ & $33.6$ \\
 & cap + DA & $0.0$ & $83.1$ & $61.5$ & $79.4$ & $36.0$ \\
\midrule
\multirow{3}{*}{SimWAM} & stock & $22.2$ & $86.2$ & $72.8$ & $78.5$ & $37.8$ \\
 & cap & $1.6$ & $82.1$ & $68.0$ & $81.0$ & $35.3$ \\
 & cap + DA & $2.0$ & $82.3$ & $65.2$ & $81.5$ & $36.1$ \\
\midrule
\multirow{3}{*}{DriveLaW} & stock & $21.5$ & $84.7$ & $57.9$ & $67.4$ & $30.0$ \\
 & cap & $0.0$ & $81.4$ & $54.4$ & $71.2$ & $28.5$ \\
 & cap + DA & $1.3$ & $81.3$ & $54.0$ & $74.3$ & $31.6$ \\
\bottomrule
\end{tabular}
\caption{Speed cap of $7.5$\,m/s as a sampling-time objective, seed 0. Viol.\ is the share of plans above the cap in percent. Strengths: $\lambda = 0.8$ (DriveWAM), $0.001$ (SimWAM), $0.0005$ (DriveLaW).}
\label{tab:supp_speed}
\end{table}

\subsection{Stabilization components}
\label{sec:supp_stab}
Table~\ref{tab:supp_stab} ablates the stabilization components on the $40$ DriveWAM failure scenes of the window study with the privileged field.

\textbf{Window}
The uniform window moves the plan further from the stock plan at the operating strength and fails earlier.

\textbf{Bound}
The bound is not active up to a strength of $16$, four times the operating strength, so removing it changes nothing there.
At strengths of $64$ and above, plans without the bound fail more often and move further.
The largest ratio of correction to model velocity at the operating points is $0.36$ on DriveWAM and $0.014$ on SimWAM and DriveLaW.

\begin{table}[t]
\centering
\scriptsize
\setlength{\tabcolsep}{3.5pt}
\begin{tabular}{lrrrrrr}
\toprule
 & \multicolumn{6}{c}{strength $\lambda$} \\
\cmidrule(lr){2-7}
Failure rate [\%] & $4$ & $8$ & $16$ & $32$ & $64$ & $128$ \\
\midrule
full method & $0.0$ & $5.0$ & $7.5$ & $20.0$ & $42.5$ & $70.0$ \\
uniform window & $5.0$ & $5.0$ & $10.0$ & $20.0$ & $37.5$ & $52.5$ \\
no bound & $0.0$ & $5.0$ & $7.5$ & $22.5$ & $60.0$ & $72.5$ \\
\midrule
FDE to stock [m] & $4$ & $8$ & $16$ & $32$ & $64$ & $128$ \\
\midrule
full method & $0.52$ & $0.63$ & $0.62$ & $0.68$ & $1.91$ & $25.1$ \\
uniform window & $0.95$ & $0.92$ & $0.74$ & $0.88$ & $1.29$ & $2.15$ \\
no bound & $0.52$ & $0.63$ & $0.62$ & $0.65$ & $2.79$ & $56.8$ \\
\midrule
Bound active [\% of steps] & $4$ & $8$ & $16$ & $32$ & $64$ & $128$ \\
\midrule
full method & $0.0$ & $0.0$ & $0.0$ & $2.8$ & $9.7$ & $25.0$ \\
\bottomrule
\end{tabular}
\caption{Full strength sweep of the stabilization study of Table~\ref{tab:stab} on $40$ DriveWAM DAC failure scenes with privileged field. $\lambda = 4$.}
\label{tab:supp_stab}
\end{table}

\subsection{Sensitivity to errors of the road-edge field}
\label{sec:supp_sens}
Table~\ref{tab:supp_sens} perturbs the predicted field on SimWAM.
The edge is shifted inward or outward by up to $1$\,m.
The score stays between $39.3$ and $39.8$ against $40.0$ for the unperturbed field, and above the stock score of $37.8$ in every case.

\begin{table}[t]
\centering
\scriptsize
\setlength{\tabcolsep}{5pt}
\begin{tabular}{lrrr}
\toprule
Field & EPDMS$\uparrow$ & DAC$\uparrow$ & EC$\uparrow$ \\
\midrule
stock, no guidance & $37.8$ & $78.5$ & $72.8$ \\
unperturbed & $40.0$ & $79.9$ & $70.1$ \\
\midrule
edge $0.25$\,m inward & $39.5$ & $79.8$ & $69.8$ \\
edge $0.25$\,m outward & $39.8$ & $79.9$ & $69.5$ \\
edge $0.5$\,m inward & $39.7$ & $80.1$ & $71.3$ \\
edge $0.5$\,m outward & $39.6$ & $79.4$ & $72.1$ \\
edge $1.0$\,m inward & $39.8$ & $80.2$ & $70.7$ \\
edge $1.0$\,m outward & $39.3$ & $79.5$ & $72.0$ \\
\bottomrule
\end{tabular}
\caption{Perturbed road-edge field on SimWAM. $\lambda = 0.05$, $\delta = 1.5$\,m.}
\label{tab:supp_sens}
\end{table}

\section{Additional qualitative results}
\label{sec:supp_qual}

\begin{figure*}[p]
\centering
\includegraphics[width=0.8\textwidth]{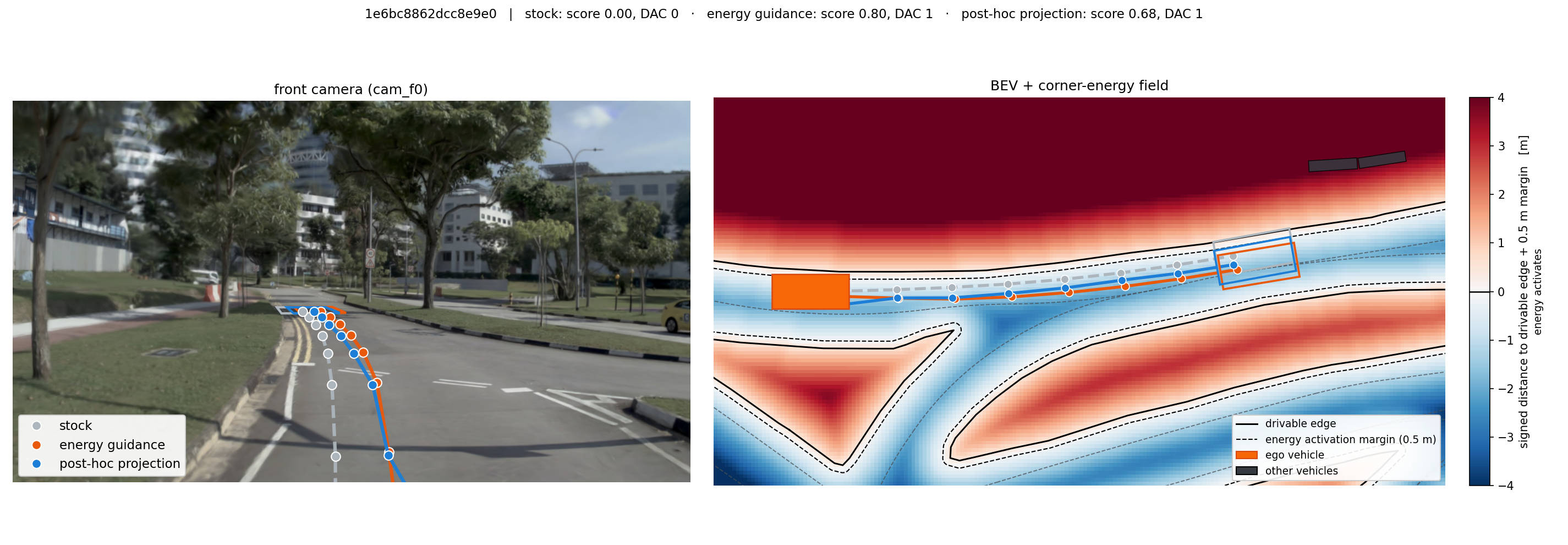}\\[2pt]
\includegraphics[width=0.8\textwidth]{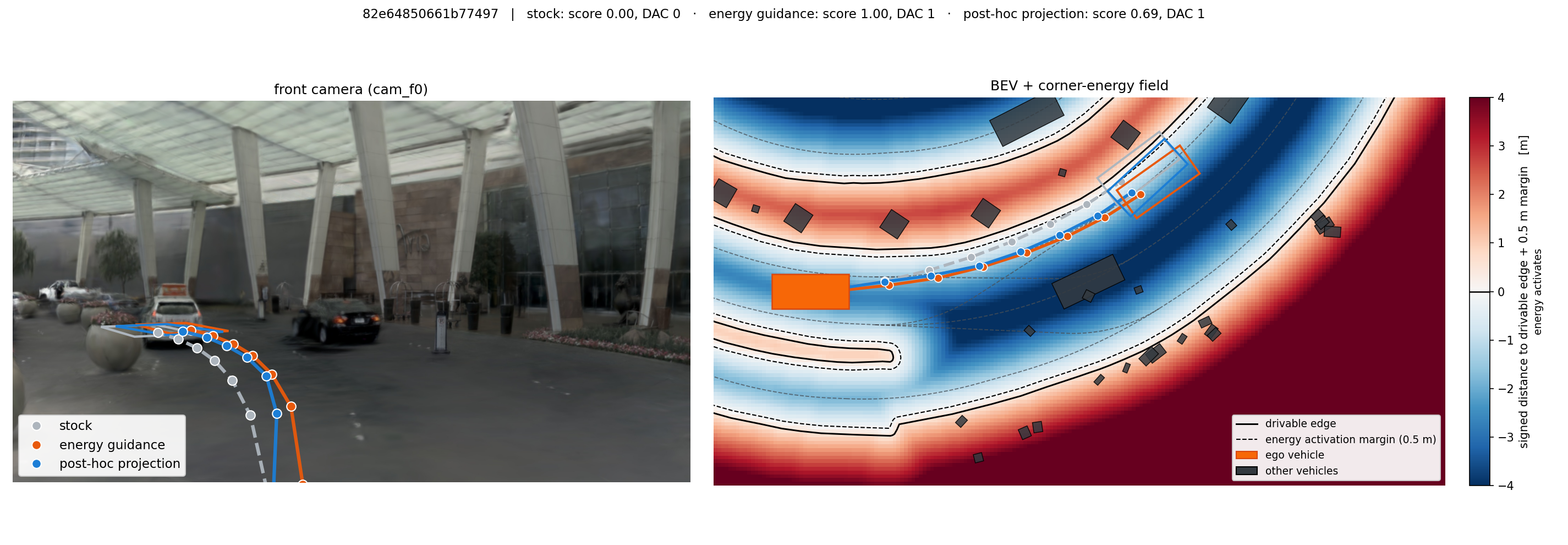}\\[2pt]
\includegraphics[width=0.8\textwidth]{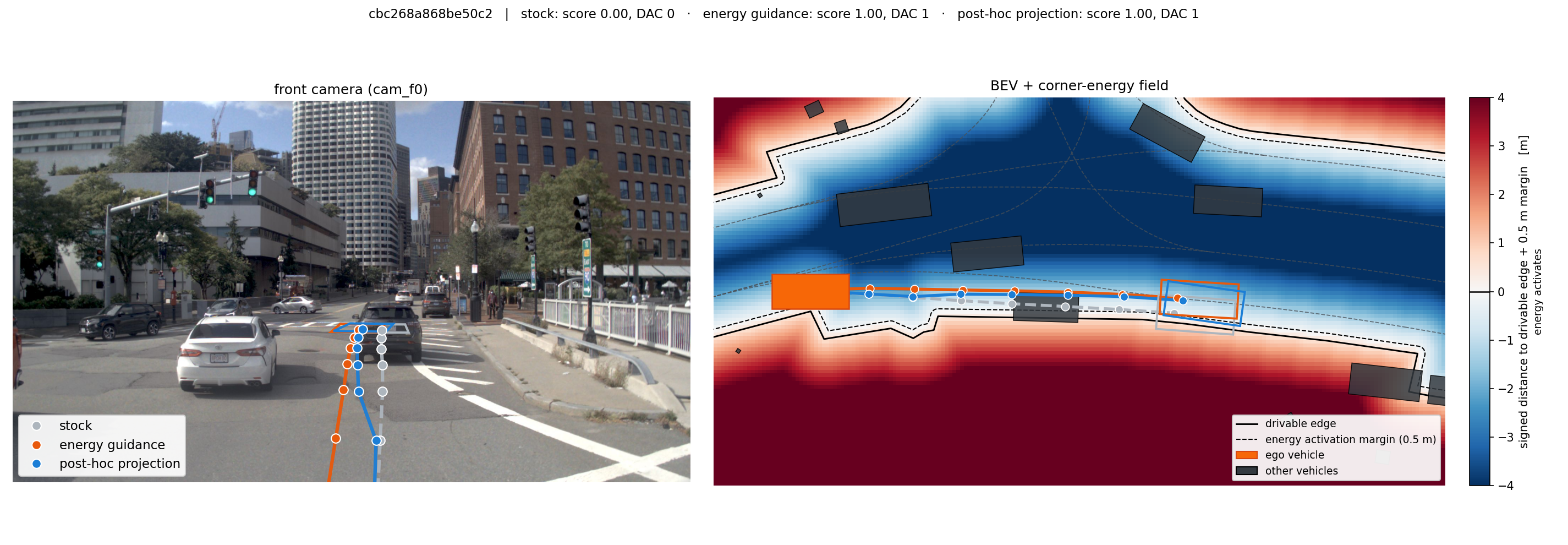}\\[2pt]
\includegraphics[width=0.8\textwidth]{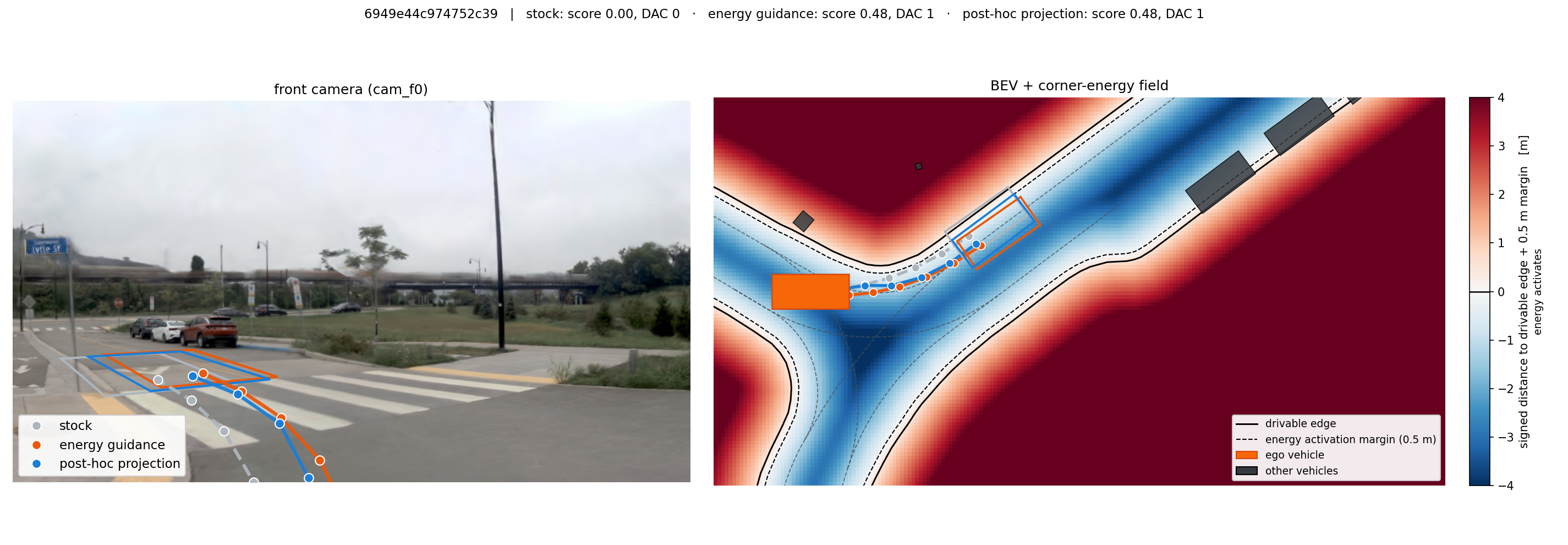}
\caption{Stock (grey), guided (orange) and projected (blue) DriveWAM plans. \textbf{Left}: front camera with the plans projected into the image. \textbf{Right}: bird's-eye view with the drivable-area energy field, the drivable edge (solid) and the activation margin (dashed).}
\label{fig:supp_qual_a}
\end{figure*}

\begin{figure*}[p]
\centering
\includegraphics[width=0.8\textwidth]{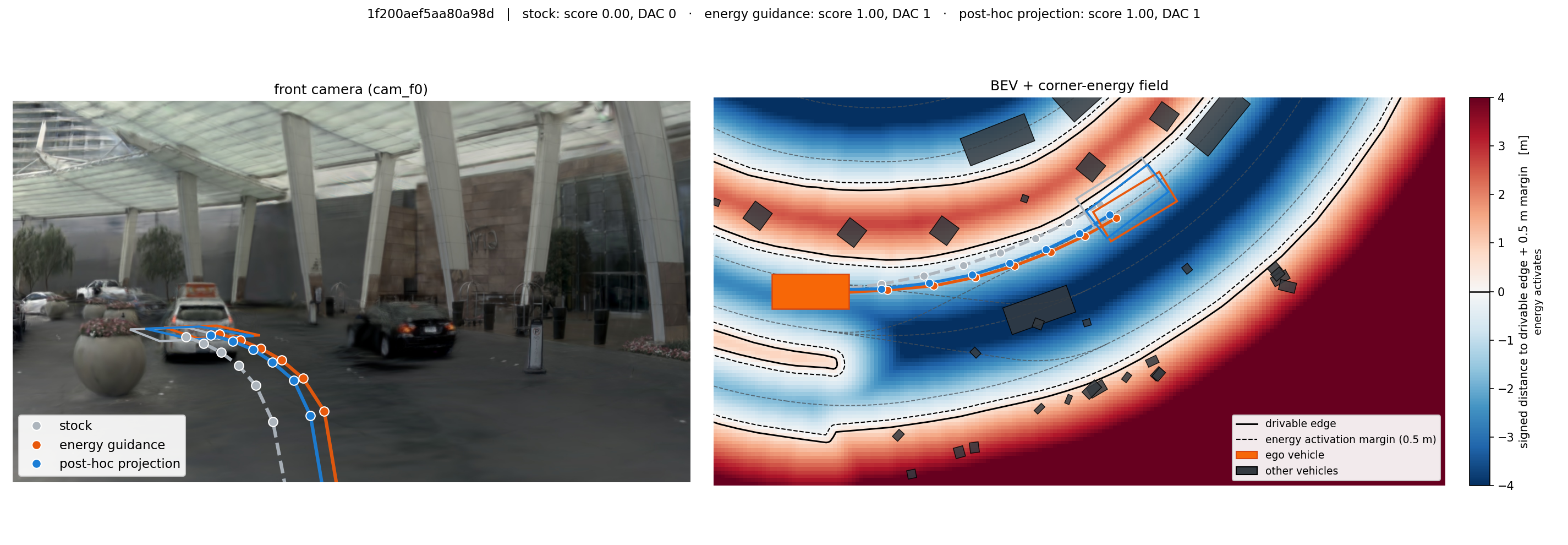}\\[2pt]
\includegraphics[width=0.8\textwidth]{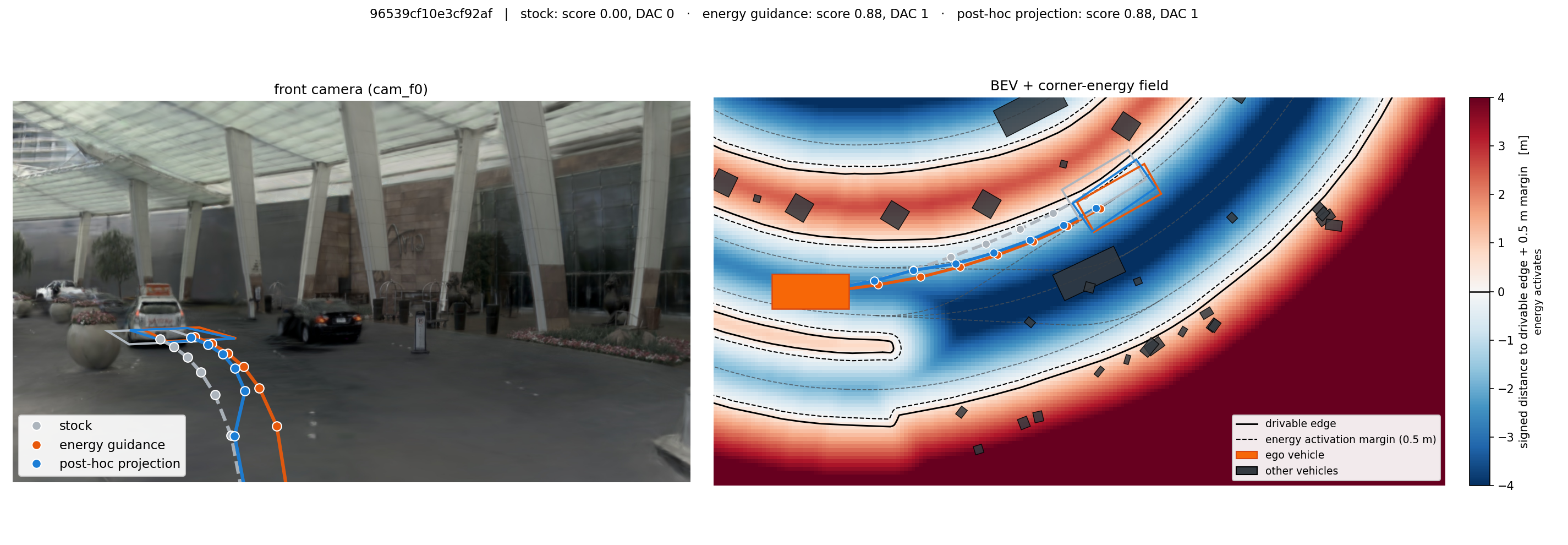}\\[2pt]
\includegraphics[width=0.8\textwidth]{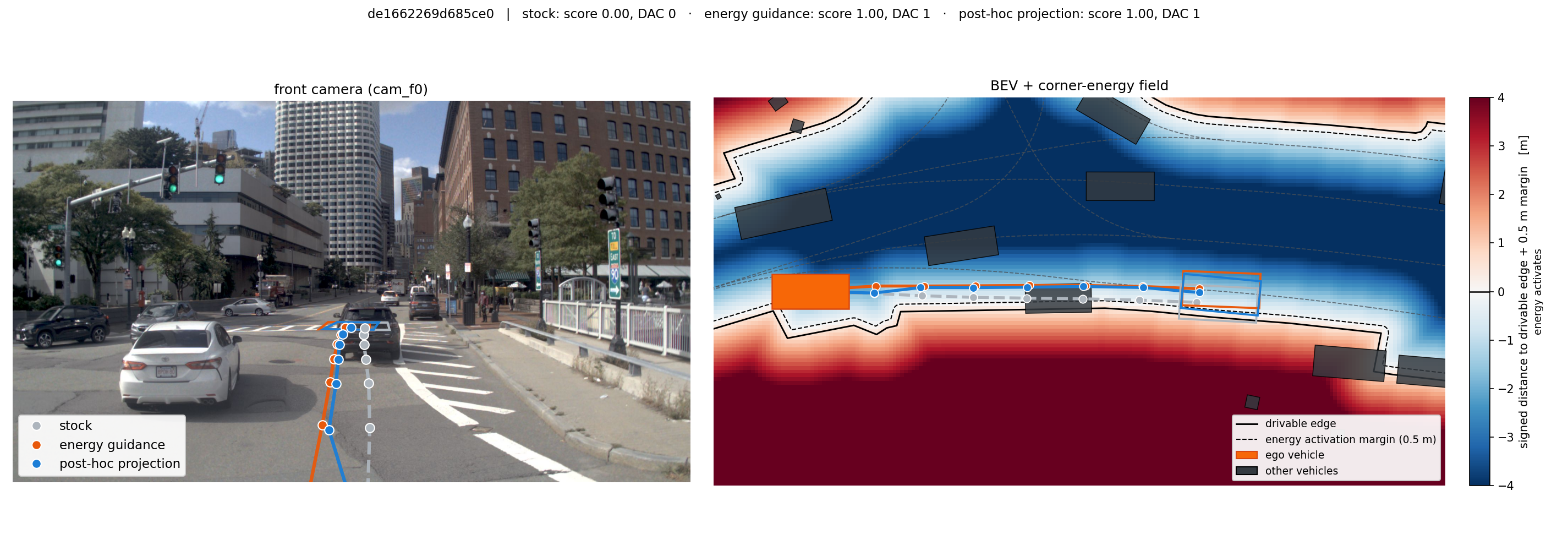}
\caption{Further scenes, same layout as Figure~\ref{fig:supp_qual_a}. The projection tends to change the plan at the waypoint of the violation, whereas the guidance shifts the whole plan.}
\label{fig:supp_qual_b}
\end{figure*}

Figures~\ref{fig:supp_qual_a} and~\ref{fig:supp_qual_b} show seven further scenes in the layout of Figure~\ref{fig:case_projection}.
On each of them the stock DriveWAM plan fails the drivable-area compliance, and both guidance and the post-hoc projection recover it.
The scenes are selected from the $109$ scenes of the hard split with this outcome, and several are synthetic variants of the same real scene.
Both methods use the privileged field with $\delta = 0.5$\,m, and guidance uses $\lambda = 4$.
The header of each panel gives the scene token and the score and drivable-area compliance (DAC) of the three plans on that scene.

